\documentclass[runningheads]{llncs}

\usepackage[mobile]{eccv}

\usepackage{eccvabbrv}

\usepackage{graphicx}
\usepackage{booktabs}
\usepackage{bbding}

\usepackage[accsupp]{axessibility}  

\usepackage[pagebackref,breaklinks,colorlinks,citecolor=eccvblue]{hyperref}

\begin{document}

\title{ConRF: Zero-shot Stylization of 3D Scenes with Conditioned Radiation Fields} 

\titlerunning{Abbreviated paper title}

\author{Xingyu Miao\inst{1} \and
Yang Bai\inst{2} \and
Haoran Duan\inst{1}\and Fan Wan\inst{1}\and Yawen Huang\inst{3}\and \\Yang Long\inst{1}\and Yefeng Zheng\inst{3}}

\authorrunning{F.~Author et al.}

\institute{Durham university, UK\and
IHPC, A*STAR, Singapore\and
Tencent Jarvis Research Center, China\\
\url{https://xingy038.github.io/ConRF/}}

\maketitle

\begin{abstract}
  Most of the existing works on arbitrary 3D NeRF style transfer required retraining on each single style condition. This work aims to achieve zero-shot controlled stylization in 3D scenes utilizing text or visual input as conditioning factors. We introduce ConRF, a novel method of zero-shot stylization. Specifically, due to the ambiguity of CLIP features, we employ a conversion process that maps the CLIP feature space to the style space of a pre-trained VGG network and then refine the CLIP multi-modal knowledge into a style transfer neural radiation field. Additionally, we use a 3D volumetric representation to perform local style transfer. By combining these operations, ConRF offers the capability to utilize either text or images as references, resulting in the generation of sequences with novel views enhanced by global or local stylization. Our experiment demonstrates that ConRF outperforms other existing methods for 3D scene and single-text stylization in terms of visual quality.
  \keywords{NeRF, Style transfer, VGG}
\end{abstract}

\section{Introduction}
\label{sec:intro}
The utilization of 3D implicit neural radiation fields has led to significant progress in generating realistic scene representations\cite{mildenhall2021nerf}, and one of the ongoing challenges in this field pertains to the application of various artistic styles in controlling the representations. Style transfer for 2D images based on neural networks has been extensively studied \cite{gatys2016image, johnson2016perceptual, gatys2015neural, ulyanov2016texture, luan2017deep, risser2017stable}, and state-of-the-art methods enable zero-shot arbitrary style transfer \cite{huang2017arbitrary,park2019arbitrary,wu2022ccpl, deng2022stytr2}. Most recently, style transfer for 3D modality has attracted increasing research attention, where the related style is changing with the shift of 2D image statistics \cite{huang2021learning, huang2022stylizednerf, mu20223d}. Existing image-driven 3D scene stylization methods lie in primarily two types of style transfer methods: zero-shot arbitrary style transfer method needs to be trained once to perform any style transfer, whereas arbitrary style transfer method necessitates retraining for each new style. Zero-shot 3D style transfer has been more popular since it only requires side information as the style like reference (e.g., images or texts) without retraining for each new style.

Several works have been proposed to explore the zero-shot 3D style transfer \cite{liu2023stylerf, chiang2022stylizing}. Although these methods can give continuous and beautiful 3D artistic scenes, they all need to give specific style references. Users who want to transfer a 3D scene to a specific art style usually need to spend time searching for a suitable reference image. Therefore, more recent work explores text-based style transfer \cite{kwon2022clipstyler, chentestnerf, fu2022language}, because text can more precisely convey the desired style without relying solely on the use of reference images. However, there is no recent work on 3D zero-shot style transfer using text as a reference style, and most work only supports arbitrary style transfer. CLIPstyler \cite{kwon2022clipstyler} is a pioneer in text-based style transfer, however, it can only be applied to 2D images and needs to be retrained for each text, which is inefficient and impractical. In the realm of 3D implicit neural fields, CLIP-NeRF \cite{wang2022clip} stands as a groundbreaking achievement, which can use text to control changes in 3D scenes.  While this innovative approach empowers the manipulation of 3D scenes, its capabilities are limited to the manipulation of color attributes exclusively, lacking the capacity to imbue artworks with distinctive stylistic elements. Although CLIP \cite{radford2021learning} has excellent image-text matching performance, it is not specially designed for style matching, thus directly leveraging CLIP will lack such information to a certain extent. Moreover, it is impractical to fine-tune CLIP for the style transfer task, because artistic style is a subjective feeling and there is no uniform standard.
\begin{figure}[t]
  \centering
  \includegraphics[width=\linewidth]{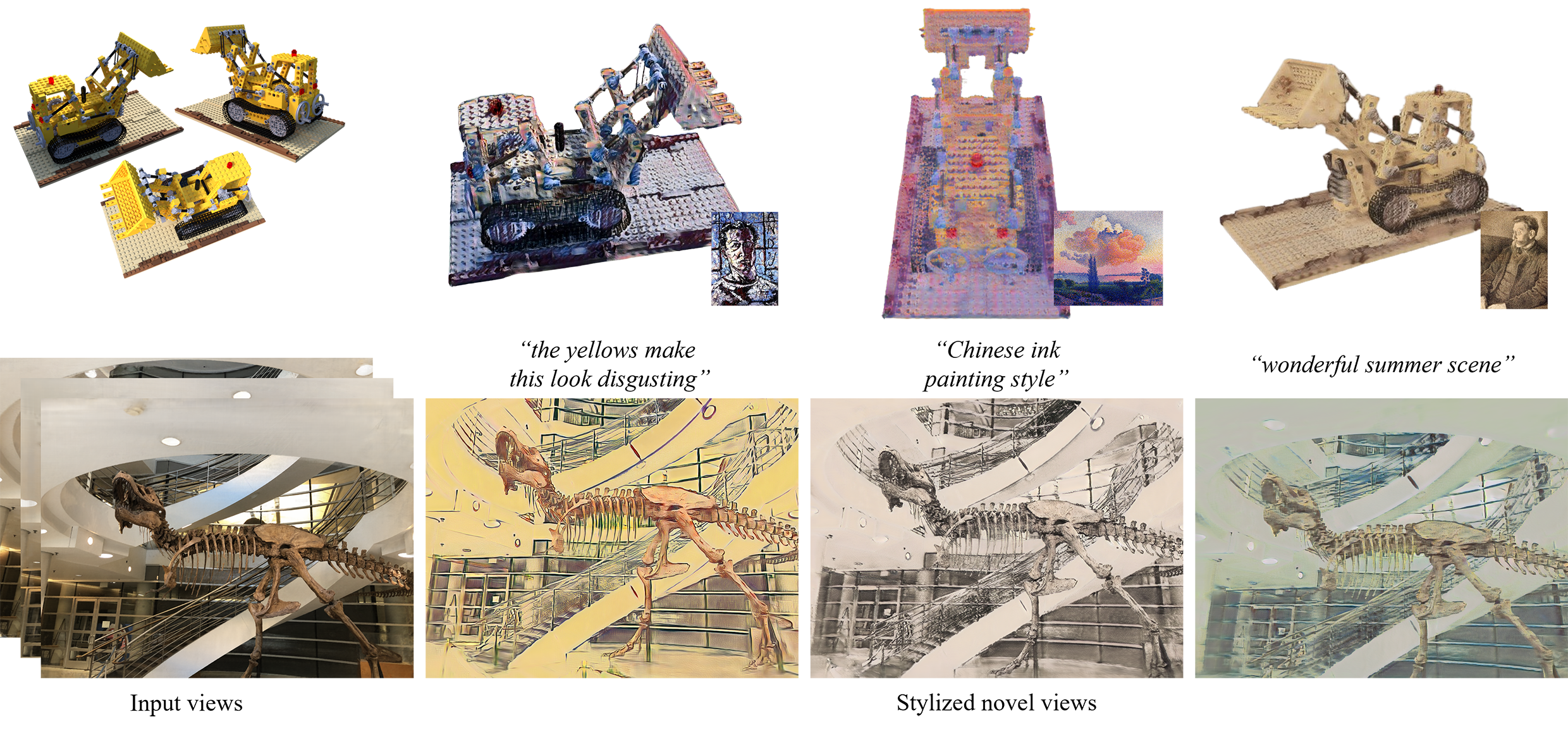}
  \caption{\textbf{Zero-shot 3D style transfer of single condition.} Given a set of multi-view content images of a 3D scene, ConRF can transfer an arbitrary text reference style or an arbitrary image reference style to the 3D scene in a zero-shot manner.}
\end{figure}

Our goal is to map the CLIP features space to the style space, simplifying the use of text or images as references to convey style. In this work, we introduce \textbf{ConRF} to achieve control over zero-shot style transfer of the neural radiation field. We leverage the encoder of CLIP to extract features, which are subsequently transformed into corresponding style features as input. These style features are then employed to transfer the style of the 3D scene. In order to achieve zero-shot transfer capabilities, we only use images to train the model in the training phase. In the inference phase, we can use text or images for inference with the help of CLIP text-image matching capabilities. To this end, we use a mapping network to map the feature embedding obtained from the CLIP encoder to the VGG feature space widely used in 2D style transfer tasks. In addition, we introduce a novel 3D selection volume to allow local style transfer to be controlled via text. In this way, our method can perform style transfer in a variety of ways, including using content text and style text or content text and style image to perform style transfer on 3D scenes. To summarize, the main contributions of this work are as follows:
\begin{itemize}
  \item
  We propose a novel method that leverages CLIP for zero-shot 3D scene artistic style transfer by a single condition (i.e. image or text).
  \item 
  We introduce a mapping network to alleviate the ambiguity in CLIP features related to style. 
  \item
  We present a 3D selection volume that allows for localized style manipulation within 3D scenes, expanding the possibilities in scene stylization and manipulation.
\end{itemize}

\section{Related Work}
\subsection{Style Transfer on Image and Video}
Artistic image stylization is a long-term problem for computer vision. The primary objective is to naturally blend the content features from one image with the style features from another, resulting in a unique image that seamlessly incorporates both elements. The early traditional method is to use handmade methods to simulate style \cite{hertzmann1998painterly, 10.1145/383259.383295}. With the development of deep learning, \cite{gatys2015neural} is the first work that uses CNN to separate and manipulate the content and style of images for style transfer. Since then, style transfer based on neural networks has gradually evolved from single-style transfer to multi-style transfer, and now to arbitrary style transfer \cite{deng2020arbitrary,huang2017arbitrary,li2019learning,li2017universal,liu2021adaattn,park2019arbitrary,sheng2018avatar,wu2021styleformer}. In terms of video style transfer, enforcing the temporal smoothness constraints defined on the optical flow can successfully transfer the style of the video \cite{chen2017coherent, huang2017real, ruder2018artistic, wang2020consistent}. Nevertheless, all stylization methods, whether for images or videos, currently rely on a predetermined reference style, imposing substantial constraints on the creative possibilities.

\subsection{Style Transfer on NeRF}
Recently, neural implicit representation methods, such as NeRF \cite{liu2020neural,mildenhall2022nerf,xian2021space,mildenhall2021nerf,xiangli2022bungeenerf,xu2022point,yu2021pixelnerf, barron2021mip, miao2024ctnerf,duan2023dynamic,miao2023ds}, have shown great potential for high-quality rendering. NeRF leverages multi-layer perceptrons (MLPs) to implicitly model continuous scenes, leading to impressive results in view synthesis when rendering scenes. The emergence of NeRF has made significant progress in the neural network representation of 3D scenes. More recently, several works \cite{chiang2022stylizing, huang2022stylizednerf, nguyen2022snerf, zhang2022arf, zhang2023transforming, liu2023stylerf, chen2022upst, fan2022unified} combine NeRF with neural style transfer \cite{gatys2016image} to handle artistic 3D scene stylization, \cite{kamata2023instruct,instructnerf2023} are based on the diffusion model and can use text to stylize 3D scenes. In particular, \cite{chiang2022stylizing, nguyen2022snerf, zhang2022arf} incorporate style loss functions from 2D images to fine-tune pre-trained NeRF models. Meanwhile, \cite{liu2023stylerf, huang2022stylizednerf} refine pre-trained NeRF models using a 2D style transfer model based on previous work. The approach presented in \cite{chen2022upst} employs a pre-trained 2D realistic network to constrain the realistic styles of different views and style images within a 3D scene. It's worth noting, however, that the primary focus of this method is on transferring only the color tone of the style image. SNeRF \cite{nguyen2022snerf} addresses the memory limitations associated with whole-image training in NeRF and enhances visual quality by employing a cyclic process of stylization and NeRF training. On the other hand, ARF \cite{zhang2022arf} and Ref-NPR \cite{zhang2023ref} aim to transfer detailed style features through matching features between style images and scenes. However, both SNeRF, ARF, and Ref-NPR entail a time-intensive optimization process for each reference style. While \cite{fan2022unified} and \cite{huang2022stylizednerf} leverage latent coding to achieve commendable results in style migration, their methods are limited to known styles and cannot adapt to unseen ones. \cite{chiang2022stylizing} can achieve arbitrary style transfer by implicitly injecting style information into MLP parameters. Similar to \cite{chen2022upst}, it can only transfer the color tone of the style image. Most of the neural style transfer-based methods mentioned above necessitate the use of style reference images to steer the style transfer, whereas our approach can leverage text or image for style transfer. To accomplish text style transfer, we utilize CLIP-encoded text embeddings to manipulate the scene for style transfer, leveraging only the already aligned text images in the pre-trained CLIP latent space. Furthermore, our work only focuses on artistic style transfer and does not have the creative capabilities based on generative models.

\section{Method}
\begin{figure}[t]
  \centering
  \includegraphics[width=0.8\linewidth]{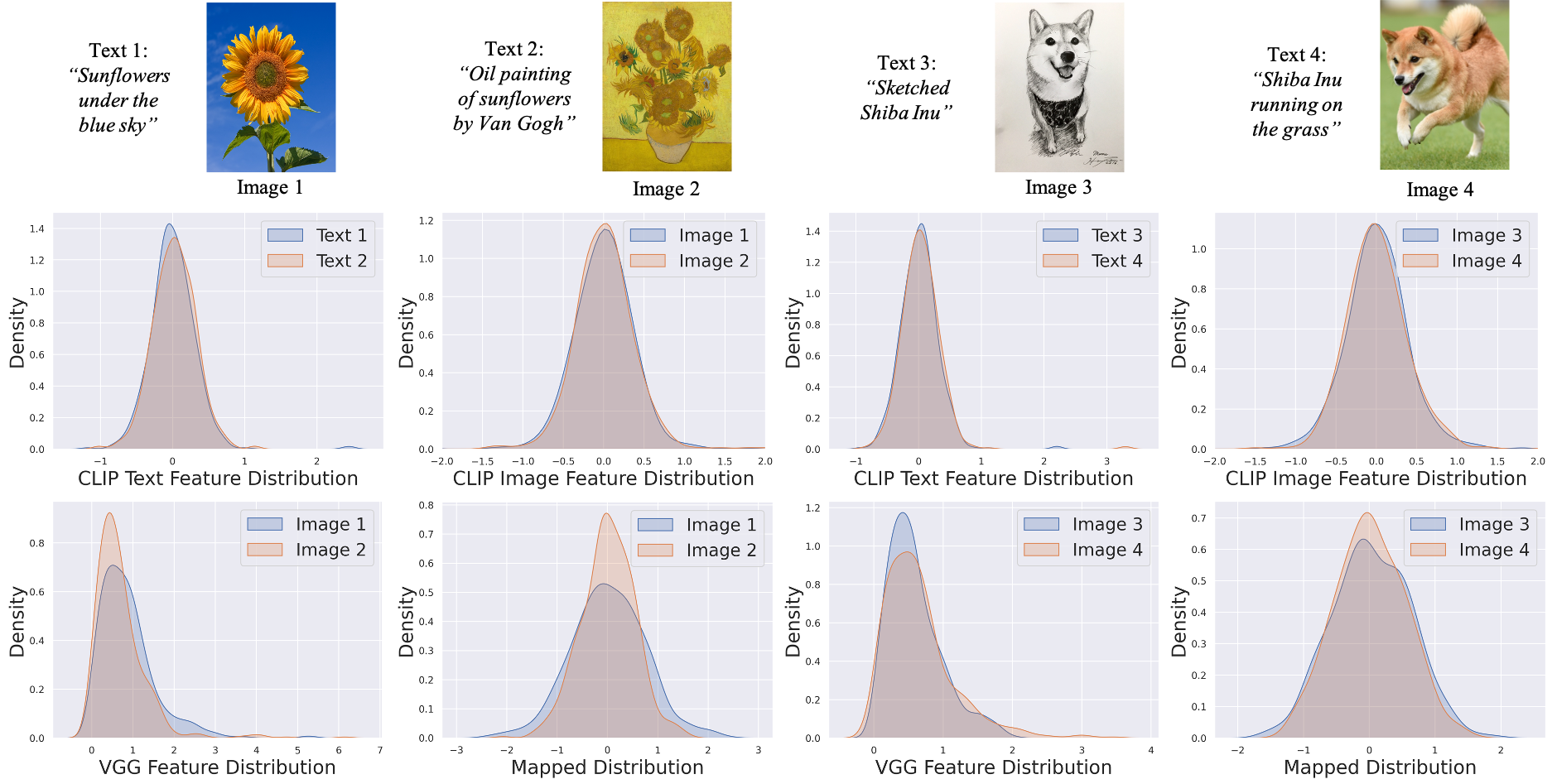}
  \caption{\textbf{Mitigating the ambiguity of CLIP features via mapping module.} The feature obtained from the CLIP extractor shows the clear high-level expression resulting in highly similar distributions for similar sunflowers or Shiba Inu with different styles, which leads to the lack of fine level features (eg. textures). On the contrary, the VGG features can better reveal the differences between the same sunflower or Shiba Inu images with different styles. To alleviate this problem, we use a mapping module to map CLIP's feature space into a style space. Features in the style space reduce this ambiguity and encourage differentiation among similar feature distributions. }
  \label{fig:motivation}
\end{figure}

In this section, we introduce ConRF, a neural radiant field method for image-text style transfer. Our goal is to utilize text and images to infuse the overall style into a 3D scene effectively. ConRF achieves this through pre-trained CLIP. However, the text-image feature space of CLIP usually has ambiguity, as illustrated in \cref{fig:motivation}. To alleviate this problem, we propose to leverage the feature space of the widely adopted pre-trained VGG model \cite{simonyan2014very} in style transfer tasks as a priori knowledge to facilitate the transformation of the CLIP feature space into a style feature space. This approach enables us to utilize pre-trained CLIP directly for style information extraction.

\begin{figure*}[t]
  \centering
  \includegraphics[width=\linewidth]{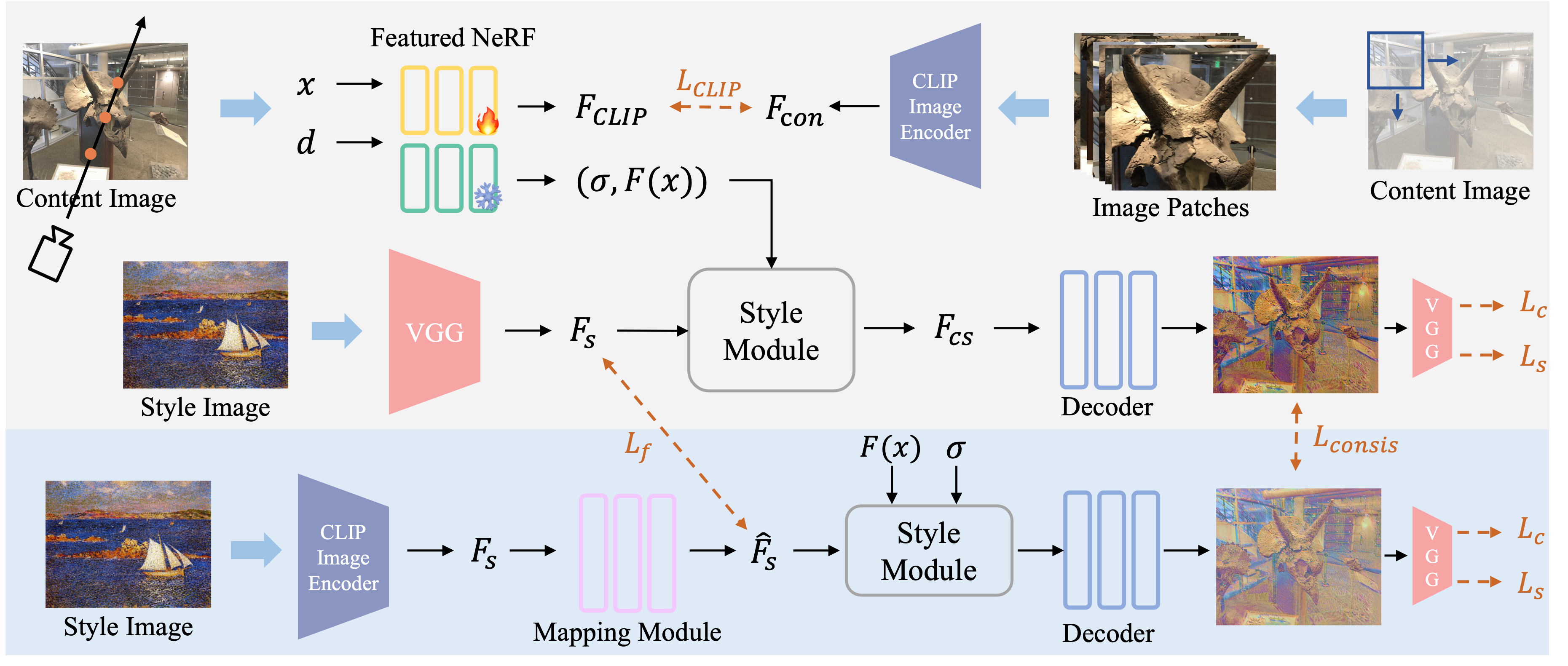}
  \caption{\textbf{The pipeline of ConRF.} ConRF performs style transfer on pre-trained feature NeRF. It consists of two branches: VGG and CLIP, which use the same style of transmission modules and share a decoder. The VGG branch uses pre-trained VGG19 \cite{simonyan2014very} to extract style features, and weakly supervises the CLIP branch using a CLIP image encoder to extract features to optimize the mapping module of the CLIP branch to separate style features. Finally, these two branches jointly optimize the decoder to obtain a stylized image. Additionally, to achieve the purpose of local transmission, we optimize an additional branch for the featured NeRF.}
  \label{fig:training}
\end{figure*}

In addition, we propose a multi-spatial feature extraction strategy to achieve local 3D scene style transfer. This strategy enhances the adaptability of the 3D CLIP volume's image-level function to pixel-level queries, ensuring precise text-to-3D scene midpoint matching. The training process of ConRF is shown in \cref{fig:training}, and the inference process is shown in \cref{fig:inference}.

\subsection{Preliminaries}
Following previous work \cite{liu2023stylerf}, we leverage the featured NeRF model to represent 3D scenes. In contrast to the original NeRF framework \cite{mildenhall2021nerf}, for every queried 3D position $x \in \mathbb{R}^3$, not only an RGB color $c$ but also furnishes a volume density $\sigma(x)$ and a multi-channel feature vector $F(x) \in \mathbb{R}^C$, where $C$ signifies the number of feature channels. Subsequently, we compute the feature representation for any rays $\textbf{r}$ intersecting the volume by performing integration over sampled points along the ray, utilizing an approximated volume rendering technique \cite{mildenhall2021nerf}:
\begin{equation}
F(\textbf{r}) = \sum_{i=1}^{N} w_iF_i,
\label{eq:preliminaries}
\end{equation}
\begin{equation}
\text{where} \quad w_i=\text{exp}\left ( -\sum_{j=1}^{i-1}\sigma_j\delta _j \right ) (1-\text{exp}(-\sigma_i\delta _i)),
\label{eq:preliminaries1}
\end{equation}
where $\sigma_i$ and $\delta_i$ refer to the volume density and feature attributes of the sampled point $i$, with $w_i$ indicating the weighting of $F_i$ within the ray $\textbf{r}$, and $\delta_i$ denoting the distance between adjacent samples. In addition, we use the same style transfer module to transfer the content style of pre-trained feature NeRF, which can be expressed as:
\begin{equation}
F_{cs} = F_c \times \sigma_I + w_\textbf{r} \times \mu_I,
\label{eq:preliminaries2}
\end{equation}
\begin{equation}
\text{where} \quad F_c = \sum_{i=1}^{N} w_iF_i, w_\textbf{r} = \sum_{i=1}^{N} w_i, \textbf{r} \in \mathcal{R}
\label{eq:preliminaries3}
\end{equation}
where $\sigma_I$ denotes standard-deviation and $\mu_I$ mean of style image ,$w_i$ denotes the weight assigned to sampled point $i$, $F_i$ stands for the feature associated with sample $i$, and $\mathcal{R}$ denotes the set of rays within each training batch. 

\subsection{Global stylization}
CLIP \cite{radford2021learning} represents a recent breakthrough in the realm of language-image pre-training. It employs two encoders, extensively trained on textual language, to construct a text-image embedding space. This space serves as a bridge connecting textual and image features, encompassing a diverse array of visual concepts. Given its ability to establish meaningful connections between text and image attributes, a natural question arises: Can this feature space be harnessed to facilitate text-based style transfer? On the one hand, because CLIP is not specifically designed for style transfer tasks, it poses challenges in directly decoupling content and style features. On the other hand, directly fine-tuning CLIP for style transfer tasks is impractical due to the highly subjective nature of style, and the absence of corresponding fine-tuning data. To this end, we propose a novel method to project text-image space onto a subspace of style features to facilitate the effective exchange of text or images with style features, so as to achieve the purpose of using text or image to control style transfer.
\paragraph{Project CLIP space to style space} It is challenging to create a text representation that matches styles in the text image space of CLIP because of insufficient data. Therefore, constructing a direct mapping from text style representation to style features through text input is not an effective method. However, features of text and images are shared in text-image space, we can learn such a mapping by using the style representation of images. To this end, we leverage a mapping module denoted as $f_\theta$ to facilitate the process. Initially, we utilize the CLIP image encoder $E_\mathcal{I}$ to extract the feature vector $F_s$ from the style image $I_s$:
\begin{equation}
F_s = E_\mathcal{I}(I_s).
\label{eq:1}
\end{equation}

Subsequently, we utilize the mapping module $f_\theta$ to establish a correspondence between the CLIP text-image feature vectors $F_s$ and the style feature representation $\hat{F}_s$:
\begin{equation}
\hat{F}_s = f_\theta(F_s),
\label{eq:2}
\end{equation}
following prior research \cite{liu2023stylerf, wu2022ccpl}, styles are typically characterized by mean ($\mu_I$) and standard-deviation ($\sigma_I$), which can be expressed as:
\begin{equation}
(\sigma_I, \mu_I) = \hat{F}_s.
\label{eq:3}
\end{equation}

Consequently, the content style can be transferred utilizing the style module, in accordance with \cref{eq:preliminaries3}.

\begin{figure*}[t]
  \centering
  \includegraphics[width=\linewidth]{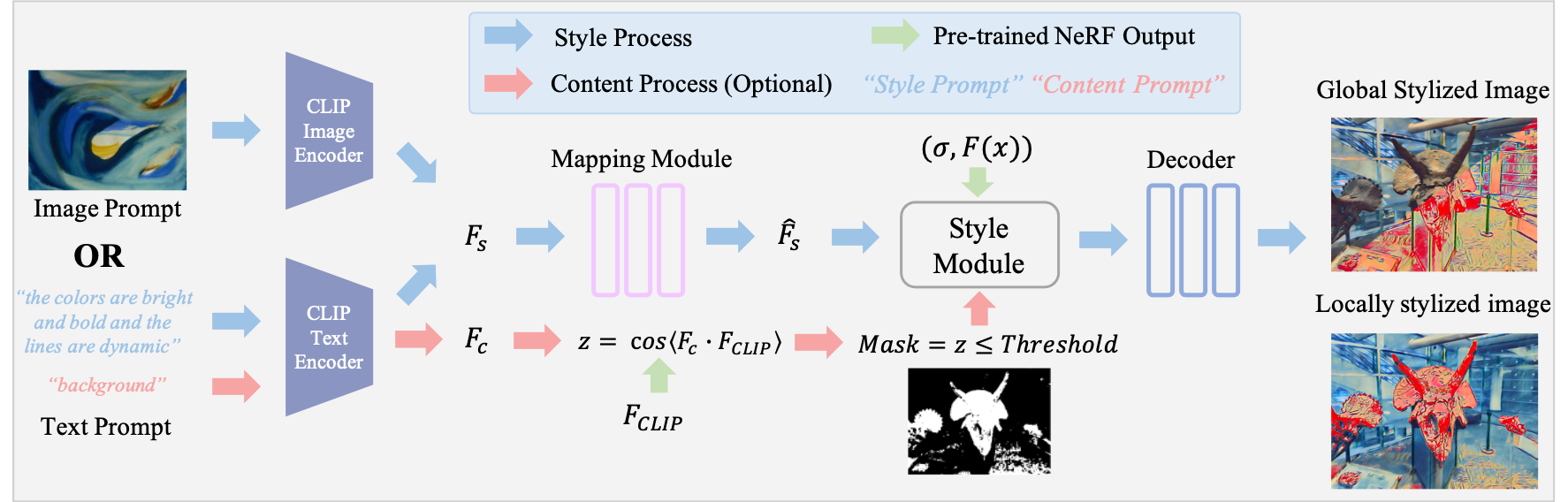}
  \caption{\textbf{The inference of ConRF.} After the training phase, ConRF is equipped to apply 3D stylistic transformations directly using text or images. Additionally, users can input specific content selection prompts to control the stylized region.}
  \label{fig:inference}
\end{figure*}

\paragraph{Weakly supervise for optimizing style space} In previous 2D studies, they usually utilize a pre-trained VGG \cite{simonyan2014very} encoder to extract features from style images, which is widely employed in style transfer. To enhance the capabilities of our mapping module and enable the projection of text-image space into the style space, we leverage pre-trained VGG \cite{simonyan2014very} to weakly supervise. To this end, we employ a two-branch approach to optimize the mapping module. The VGG branch takes as input the features from the style transfer module and utilizes a pre-trained VGG \cite{simonyan2014very} to extract style features. Meanwhile, the CLIP branch takes as input the features that the mapping module has projected. To train the CLIP branch, we introduce style feature loss to ensure consistency between the output of the mapping module and the style features of the VGG branch, which can be expressed as:
\begin{equation}
\mathcal{L}_f = \left \| F_s^v - \hat{F_s^c} \right \| _2^2, 
\label{eq:4}
\end{equation}
\begin{equation}
\text{where} \quad F_s^v = (\sigma_v, \mu_v), \hat{F_s^c} = (\sigma_c, \mu_c), 
\label{eq:5}
\end{equation}
where $F_s^v$ denotes that the VGG branch uses $\texttt{ReLU\_3\_1}$ layer of the pre-trained VGG \cite{simonyan2014very} extract style features form style image, $\hat F_s^c$ represents the output of the mapping module of the CLIP branch, and $\sigma, \mu$ represent the standard-deviation and mean respectively. Thus, the $\mathcal{L}_f$ is:
\begin{equation}
\mathcal{L}_f = \left \| \sigma_v - \sigma_c \right \| _2^2 + \left \| \mu_v - \mu_c \right \| _2^2. 
\label{eq:6}
\end{equation}

Furthermore, we adopt the identical loss function employed in prior studies \cite{wu2022ccpl, liu2023stylerf, huang2017arbitrary} for both the VGG and CLIP branches to train the shared weight decoder. Specifically, the content loss, denoted as $\mathcal{L}_{c}$, is computed as the Mean Squared Error (MSE) of the feature map, while the style loss, represented as $\mathcal{L}_s$, is computed as the MSE of the mean and standard deviation of the channel features:
\begin{equation}
\mathcal{L}_{stylized} = \mathcal{L}_{c} + \lambda\mathcal{L}_s, 
\label{eq:7}
\end{equation}
where $\lambda$ controls the balance between content preservation and stylization effects. It's important to note that both $\mathcal{L}_{c}$ and $\mathcal{L}_s$ incorporate from both the VGG and CLIP branches. Finally, in order to reduce the uncertainty of the optimized decoder caused by using different feature extractors for the VGG and CLIP branches, we use a consistency loss, denoted as $\mathcal{L}_{consis} = \left \| I_s^v - I_s^c \right \|_1$, where $I_s^v$ is the final stylized image of the VGG branch, and $I_s^c$ is the final stylized image of the CLIP branch.

\begin{figure*}[t]
  \centering
  \includegraphics[width=\linewidth]{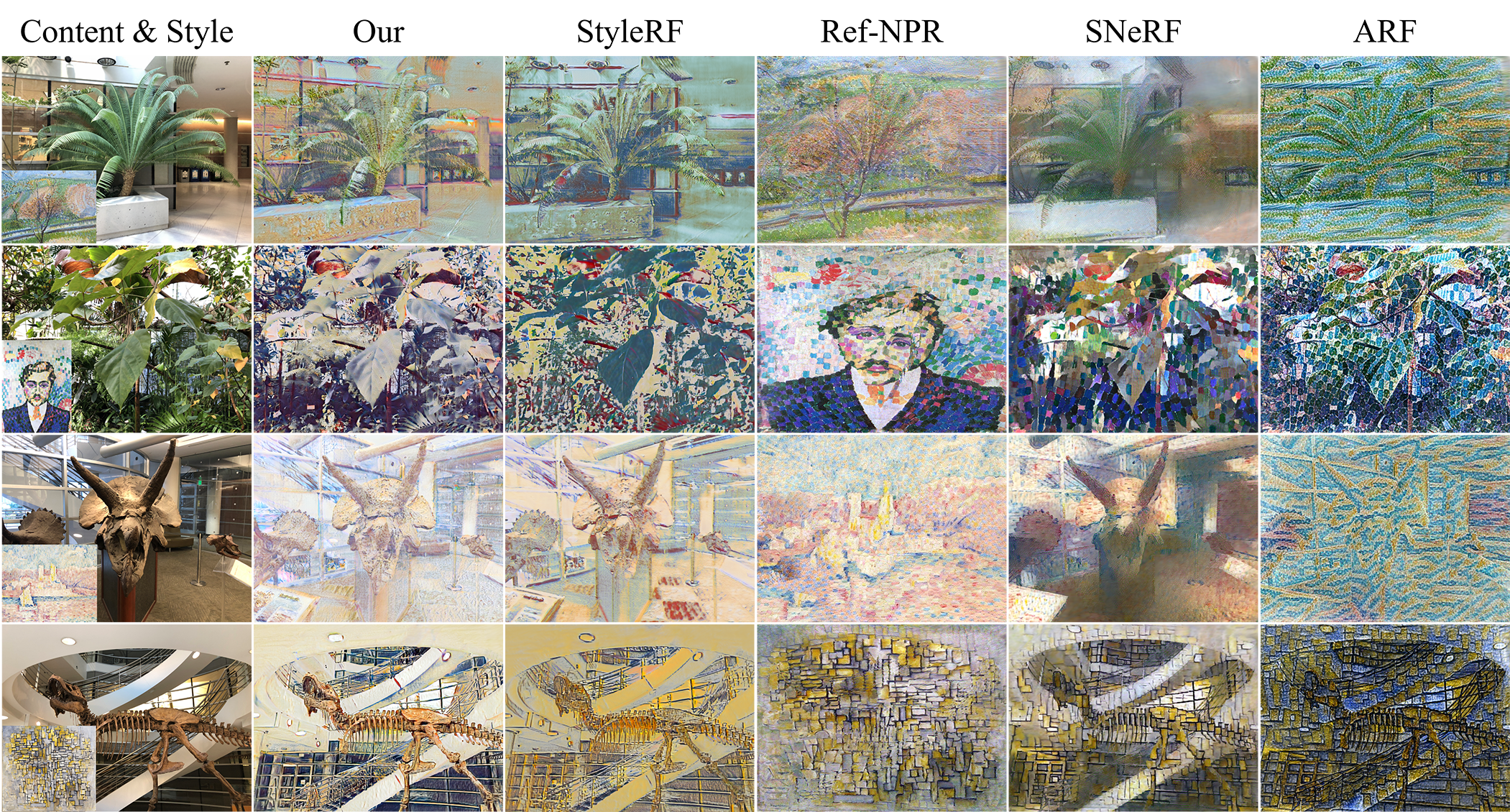}
  \caption{\textbf{Comparison with four SOTA 3D style transfer methods using reference style images.} For the four scenes in the example our method produces significantly better 3D style transfer.}
  \label{fig:qualitative}
\end{figure*}

\subsection{Local stylization}
Prior work \cite{liu2023stylerf}  allowed different styles on different objects. However, this approach necessitated the provision of precise masks in advance to regulate the styles for achieving the desired combinations. Taking inspiration from recent promising prompt-based methods, we introduce a simple approach using text prompts to apply styles directly to specific 3D regions.

\paragraph{Utilizing CLIP features weakly supervise a 3D selection volume}
Given a multi-view image depicting a scene and a selected textual content description, our objective is to execute partial style transfer on the reconstructed NeRF. This process assigns a matching style to each designated 3D point. To achieve this, we exploit the multimodal capabilities of the CLIP model to associate each 3D point with CLIP features and their corresponding semantic information. Specifically, we introduce an additional branch to get 3D CLIP feature volume for rendering the CLIP feature, we can then render the CLIP feature of each ray $\textbf{r}$ using volume rendering \cite{mildenhall2021nerf}:
\begin{equation}
F(\textbf{r})_{\text{CLIP}} = \sum_{i=1}^{N} w_iF_i.
\label{eq:8}
\end{equation}

Next, we can weakly supervise to optimize the CLIP features from NeRF rendering. We employ the CLIP image encoder to obtain content features $F_c$ and optimize them using the $L1$ Loss:
\begin{equation}
\mathcal{L}_{\text{CLIP}} = \left \| F(\textbf{r})_{\text{CLIP}} - F_{con} \right \|_1 .
\label{eq:9}
\end{equation}
However, since CLIP only generates image-level features, when using text for query, it can only match image-level features but not pixel-level features. Therefore, we segment training images into patches of varying sizes, extract CLIP features from these patches, and subsequently integrate these CLIP features to derive multi-spatial features and enable the 3D selection volume to learn the ability to be used for pixel-level feature queries.

\paragraph{Multi-spatial feature extraction strategy}
We suggest an innovative approach that relies on multi-spatial strategies to enhance the adaptability of CLIP's image-level features for pixel-level queries. The multi-spatial component is designed to extract features from patches, each containing pixels located at distinct positions. To achieve this, we employ the sliding window algorithm for multi-spatial feature extraction and then calculate the average of these multi-spatial features. First, we initialize two tensors $\mathcal{C}$ and $\mathcal{F}$ to record the counts and features within the sliding window. We adopt a sliding window approach to partition the input image. Within each window, we extract image data and use a CLIP image encoder to obtain features, denoted as $E_\mathcal{I}(x, y)$, where $x$ and $y$ represent pixel positions. The features and counts for each window are cumulatively updated in the $\mathcal{C}$ and $\mathcal{F}$ via:
\begin{equation}
\mathcal{C}(x,y) += 1,
\label{eq:10}
\end{equation}
\begin{equation}
\mathcal{F}(x,y) += E_\mathcal{I}(x, y).
\label{eq:11}
\end{equation}
Finally, the normalized feature representation is obtained, which is expressed as:
\begin{equation}
\mathcal{F}_{normal}(x,y) = \frac{\mathcal{F}(x,y)}{\mathcal{C}(x,y)},
\label{eq:12}
\end{equation}
where $\mathcal{F}_{normal}$ encapsulates the comprehensive average feature representation of the entire image, integrating information from all windows.

\begin{figure*}[t]
  \centering
  \includegraphics[width=\linewidth]{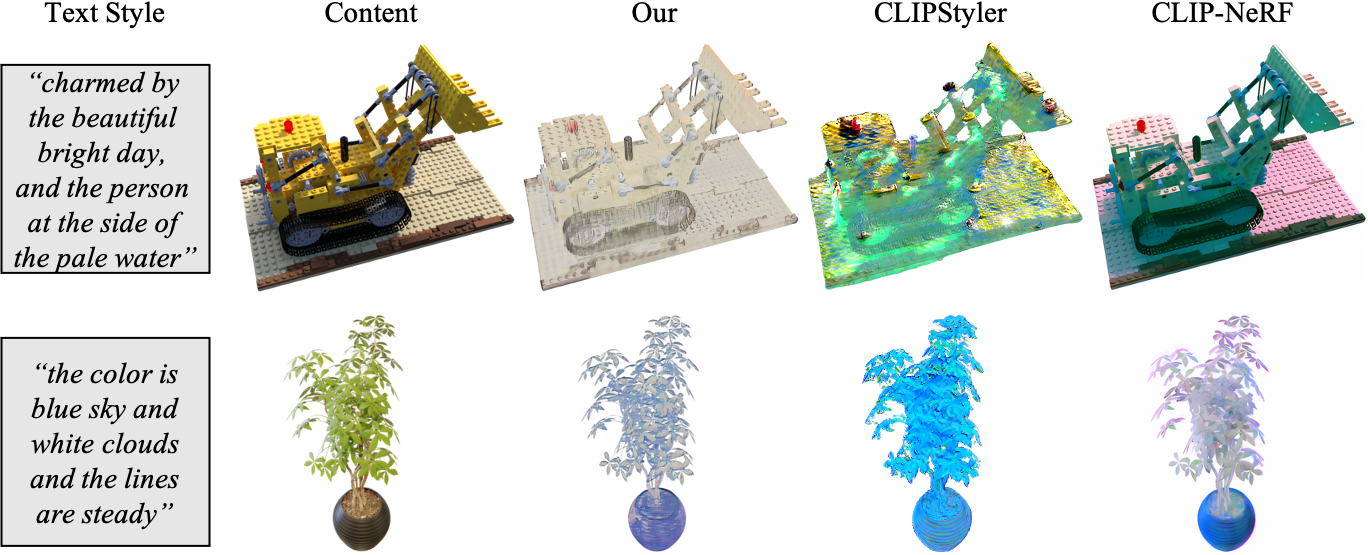}
  \caption{\textbf{Comparison with two SOTA style transfer methods using text prompt on Synthetic NeRF.} For the two scenes in the example, our approach produces 3D style transfers that are significantly closer to textual descriptions.}
  \label{fig:text_qualitative}
\end{figure*}

\subsection{Inference process}
As shown in \cref{fig:inference}, once training is completed, we can use text to perform style transfer on 3D scenes. Specifically, since text and image features are shared,  we can get the mean and variance of the style when we input text, that is,
\begin{equation}
(\sigma_T, \mu_T) = f_\theta(E_\mathcal{T}(T_s)),
\label{eq:13}
\end{equation}
where $T_s$ denotes the text input, $E_\mathcal{T}$ denotes the CLIP text encoder and $(\sigma_T, \mu_T)$ is the style representation obtained from the text description. Thus, using CLIP for style transfer can be expressed as:
\begin{equation}
F_{cs} = \sum_{i=1}^{N} w_i(F_i \times \sigma(\hat{F}_s) + \mu(\hat{F}_s)).
\label{eq:14}
\end{equation}
\cref{eq:14} can be interpreted as the individual transfer of style to every sampling point along the ray prior to volume rendering. Ultimately, we employ a decoder to reproject these stylized features into RGB space, thereby yielding stylized novel views. In addition, our local style transfer feature is optional and becomes active only once a content text prompt is entered. 
The local style transfer branch is activated, we can use a content text prompt and $F_c$ calculate the similarity $z$:
\begin{equation}
z = cos\left \langle F_c, F_{\text{CLIP}}\right \rangle,
\end{equation}
where $cos\left \langle , \right \rangle $ is the cosine similarities. We can then get the mask $M$:
\begin{equation}
m = z \leq t,
\end{equation}
$t$ is the threshold. Thuse \cref{eq:14} can be replace:

\begin{equation}
\begin{split}
&F_{cs}^{\text{local}} = M\sum_{i=1}^{N} w_i(F_i \times \sigma(\hat{F}_{s1}) + \mu(\hat{F}_{s2})) \\&+ (1-M)\sum_{i=1}^{N} w_i(F_i \times \sigma(\hat{F}_{s2}) + \mu(\hat{F}_{s2})).
\end{split}
\end{equation}

Our method supports both reference image style transfer and text prompt style transfer, enabling versatile multi-combination style transfers. After inputting content text prompts, our method allows for the use of various prompt combinations to transform different aspects. It supports combinations such as text-text, image-image, and text-image for style transfers.

\begin{figure}[t]
  \centering
  \includegraphics[width=0.7\linewidth]{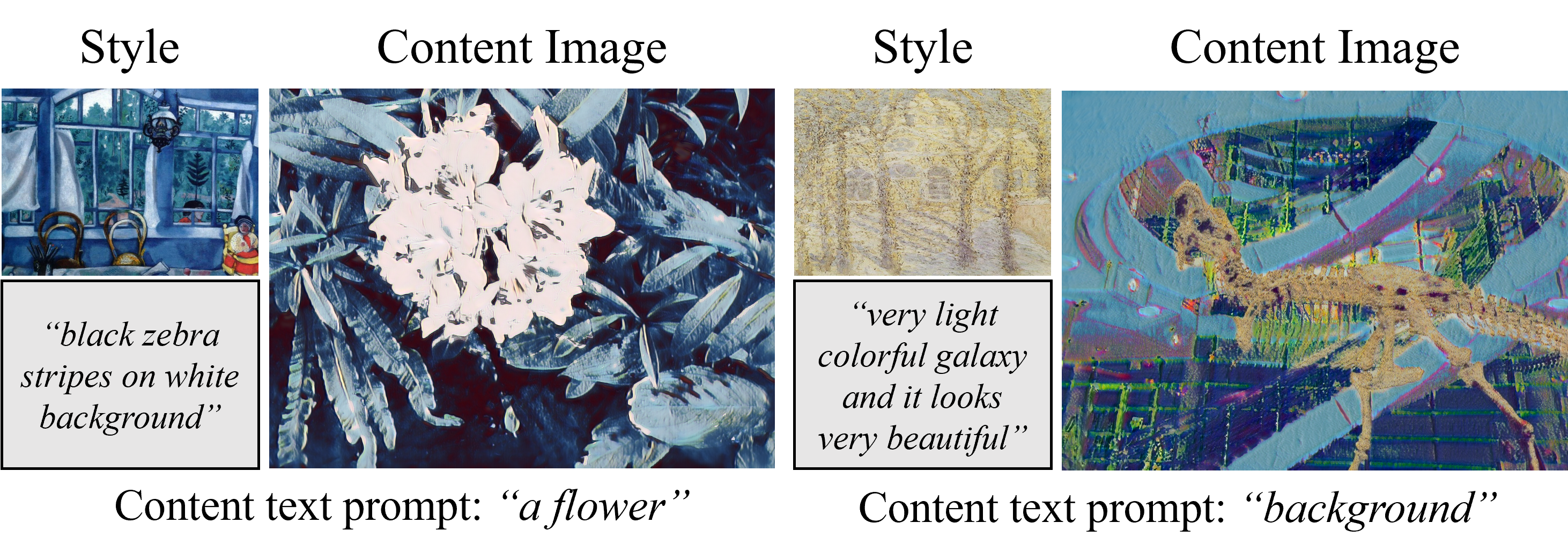}
  \caption{\textbf{Local style transfer results.}  We use the text prompt to select content, and then use images and text to perform style transfer on different areas.}
  \label{fig:local_stylized}
\end{figure}

\section{Experiment}
In this section, we assessed the proposed method through both qualitative and quantitative experiments. Given the versatility of our approach, which supports style transfer with text and image references, we conducted a comparison against the SOTA methods in these two aspects.

\subsection{Qualitative Experiments}
We conducted a comprehensive evaluation of ConRF using two publicly available datasets: LLFF \cite{mildenhall2019local}, which comprises real scenes with intricate geometry, and Synthetic NeRF \cite{mildenhall2022nerf}, featuring $360^\circ$ views of objects. For methods that rely on reference-style images, we benchmarked against the state-of-the-art StyleRF \cite{liu2023stylerf}, Ref-NPR \cite{zhang2023ref}, SNeRF \cite{nguyen2022snerf}, and ARF \cite{zhang2022arf}. In the case of methods employing text prompts, ConRF was evaluated against the 2D single-text condition method, CLIPStyler \cite{kwon2022clipstyler}, and a 3D text control editing method, CLIP-NeRF \cite{wang2022clip}.

\begin{table}
\centering
\begin{tabular}{ccccc}
\toprule
Methods & \multicolumn{2}{c}{\begin{tabular}[c]{@{}c@{}}Short-range\\ Consistency\end{tabular}} & \multicolumn{2}{c}{\begin{tabular}[c]{@{}c@{}}Long-range\\ Consistency\end{tabular}} \\ \hline
        & \textbf{SSIM}                                      & \textbf{LPIPS}                                     & \textbf{SSIM}                                      & \textbf{LPIPS}                                    \\
Ref-NPR \cite{zhang2023ref} & 0.782                                     & \textcolor{red}{0.028}                                     & 0.337                                     & \textcolor{red}{0.088}                                    \\
SNeRF \cite{nguyen2022snerf}   & \textcolor{red}{0.852}                                     & \textcolor{blue}{0.029}                                     & \textcolor{blue}{0.401}                                     & \textcolor{blue}{0.107}                                    \\
ARF \cite{zhang2022arf}    & \textcolor{blue}{0.816}                                     & 0.056                                     & 0.336                                     & 0.148                                    \\
StyleRF \cite{liu2023stylerf} & 0.495                                     & 0.059                                     & 0.050                                     & 0.217                                    \\
\textbf{Our}     & 0.763                                     & 0.041                                     & \textcolor{red}{0.442}                                     & \textcolor{red}{0.088}                                    \\ \hline 
$\text{CLIPStyler}^*$ \cite{kwon2022clipstyler}
 & 0.606                                     & 0.048                                     & 0.356                                     & 0.254                                    \\
$\text{CLIP-NeRF}^*$ \cite{wang2022clip}
   & \textcolor{blue}{0.641}                                     & \textcolor{blue}{0.043}                                     & \textcolor{blue}{0.389}                                     & \textcolor{blue}{0.162}                                    \\
$\text{\textbf{Our}}^*$     & \textcolor{red}{0.787}                                     & \textcolor{red}{0.038}                                     & \textcolor{red}{0.524}                                     & \textcolor{red}{0.091} \\
\bottomrule
\end{tabular}
\caption{\textbf{Results on consistency.}  We compare ConRF with the state-of-the-art on consistency using LPIPS ($\downarrow$) and SSIM($\uparrow$). ($*$) is using a text prompt to style transfer. The best score is \textcolor{red}{red}, and the second score is \textcolor{blue}{blue}.}
\label{tab:1}
\end{table}

The qualitative comparisons are shown in \cref{fig:qualitative} and \cref{fig:text_qualitative}. For methods using reference-style images, we can see that compared with other methods, our method transfers the color and artistic style of the reference-style image while maintaining relatively complete original content. Although the single style transfer methods SNeRF \cite{nguyen2022snerf} and ARF \cite{zhang2022arf} show better transfer results compared to our method (as shown in the Trex scenario), we transfer style features in localized regions as well and can perform arbitrary style transfers without retraining. In addition, for the Ref-NPR \cite{zhang2023ref}, which is highly dependent on the existing 2D transmission method. In order to make the comparison fair, we directly use the style image as the reference style for transfer. In terms of methods using text prompts, compared with CLIPStyler \cite{kwon2022clipstyler} and CLIP-NeRF \cite{wang2022clip}, our method is closer to text description. Although our method can achieve excellent results, we observe that our method can be further improved in some aspects, which are described in the limitations of supplementary material. We present the locally transferred image in \cref{fig:local_stylized}. Our method presents the ability to utilize text prompts effectively, allowing for direct and precise control over specific sections of content. The selection volume empowers users to effortlessly convey a wide range of distinct styles within localized areas. At present, our approach serves as an initial exploration, but it holds great promise.

\begin{figure}
  \centering
  \includegraphics[width=0.7\linewidth]{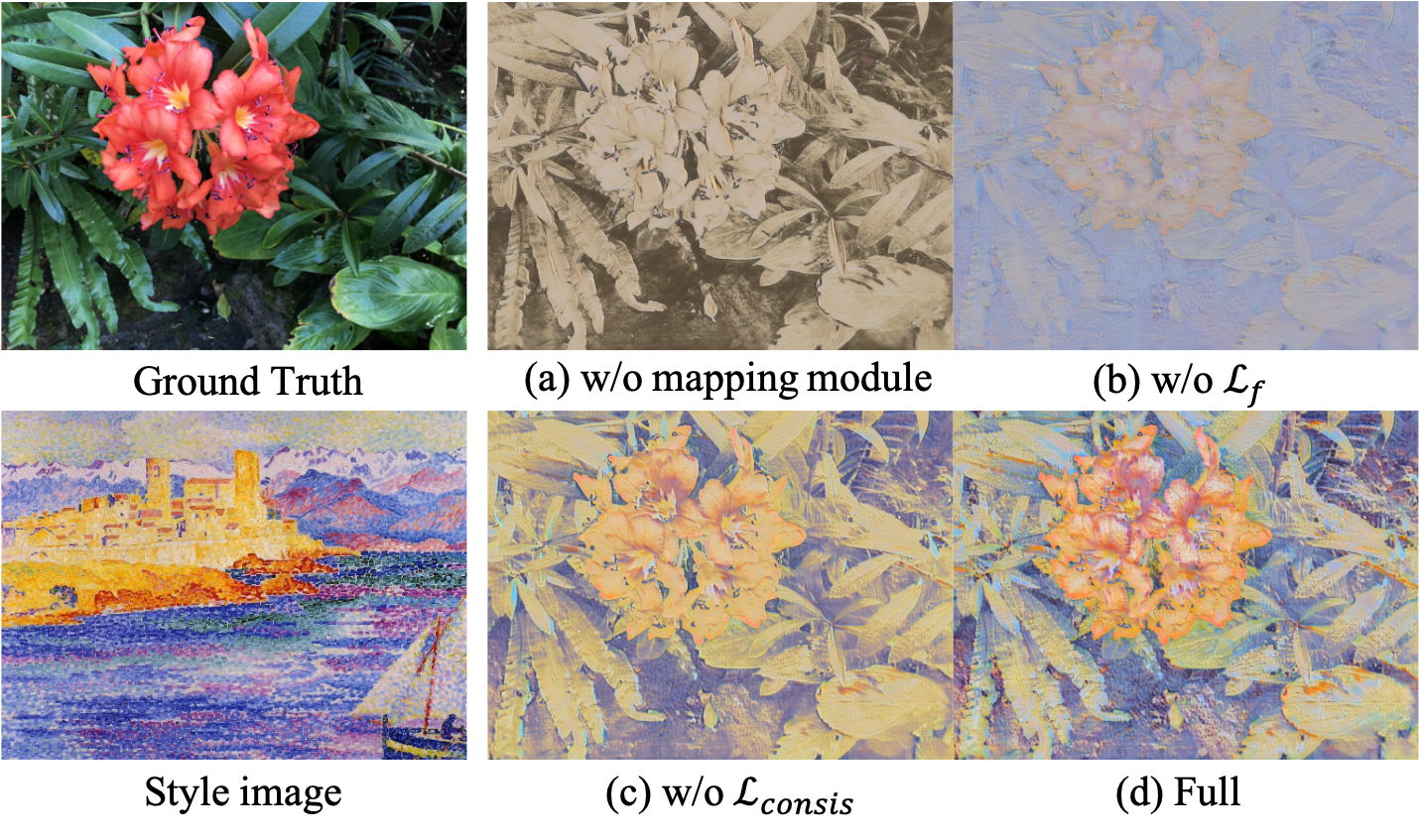}
  \caption{\textbf{Ablation studies.} (a) shows the stylization without the mapping module, (b) shows the stylization without style feature loss, (c) shows the stylization without consistency loss, and (d) shows the stylization of our full pipeline.}
  \label{fig:ab_main}
\end{figure}

\subsection{Quantitative Results}
At present, 3D style transfer stands as a nascent and scarcely explored domain, with limited metrics available for quantitatively evaluating stylization quality. Therefore, following \cite{chiang2022stylizing, chen2022upst} we compared the consistency of multiple views, we compared the short-range and long-range consistency scores of adjacent views and distant views respectively. In our experiments, we warp one view to another based on optical flow \cite{teed2020raft} and then compute masked SSIM and LPIPS scores \cite{zhang2018unreasonable} to measure the stylization consistency. 

As shown in \cref{tab:1}, we compared with six methods. Compared to our baseline StyleRF \cite{liu2023stylerf}, our short-range LPIPS score is improved by 31\%, and long-range LPIPS score is improved by 59.4\%. Please note that SNeRF \cite{nguyen2022snerf} and Ref-NPR \cite{zhang2023ref} outperform our method. This is primarily due to the fact that in certain scenarios, the style transfer effect of SNeRF \cite{nguyen2022snerf} may not be as pronounced, whereas Ref-NPR \cite{zhang2023ref} directly transfers the style image's characteristics into the content. For methods using text prompts, we are completely superior to both methods.

\begin{figure}[t]
  \centering
  \includegraphics[width=0.8\linewidth]{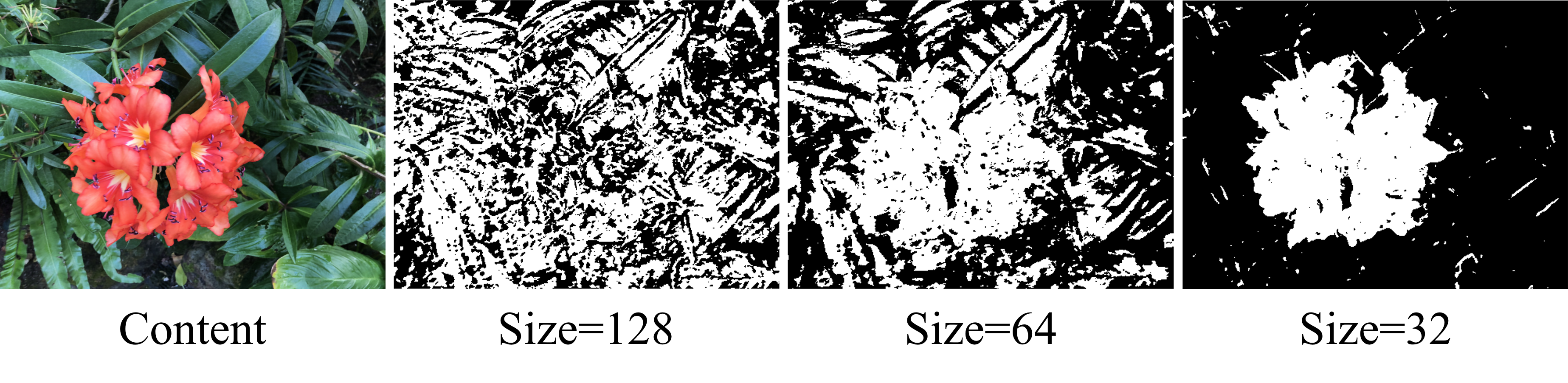}
  \caption{\textbf{Ablation studies} for the multi-spatial feature extraction strategy. The text prompt is 'a flower,' and the mask is generated using a selection volume constructed with different window sizes. When $size=32$, the mask is the cleanest.}
  \label{fig:ab_mask}
\end{figure}

\subsection{Ablation Studies}
To evaluate the effectiveness of our approach, we perform ablation experiments separately for global style transfer and local style transfer functions.

\paragraph{Golbal stylization} In \cref{fig:ab_main}, we present a comparison between our full system and its variants, where we removed specific modules individually: A) mapping module, B) style feature loss, and C) consistency loss. As illustrated, the absence of a mapping module impedes the transfer of style from any reference image. The lack of a style feature loss compromises the quality of the transferred style, even though some elements of the reference picture's style may still be obtained. Moreover, the inclusion of a consistency loss noticeably enhances the performance of style transfer.

\paragraph{Local stylization} We control local style transfer through text prompts. As illustrated in \cref{fig:ab_mask}, we conduct a comprehensive evaluation of the Multi-spatial feature extraction strategy by experimenting with different sliding window sizes. Our observations indicate that utilizing a window size of 32 leads to the generation of a notably refined mask, effectively achieving our goal of fine-grained style control in localized areas.

\begin{figure}[t]
  \centering
  \includegraphics[width=0.8\linewidth]{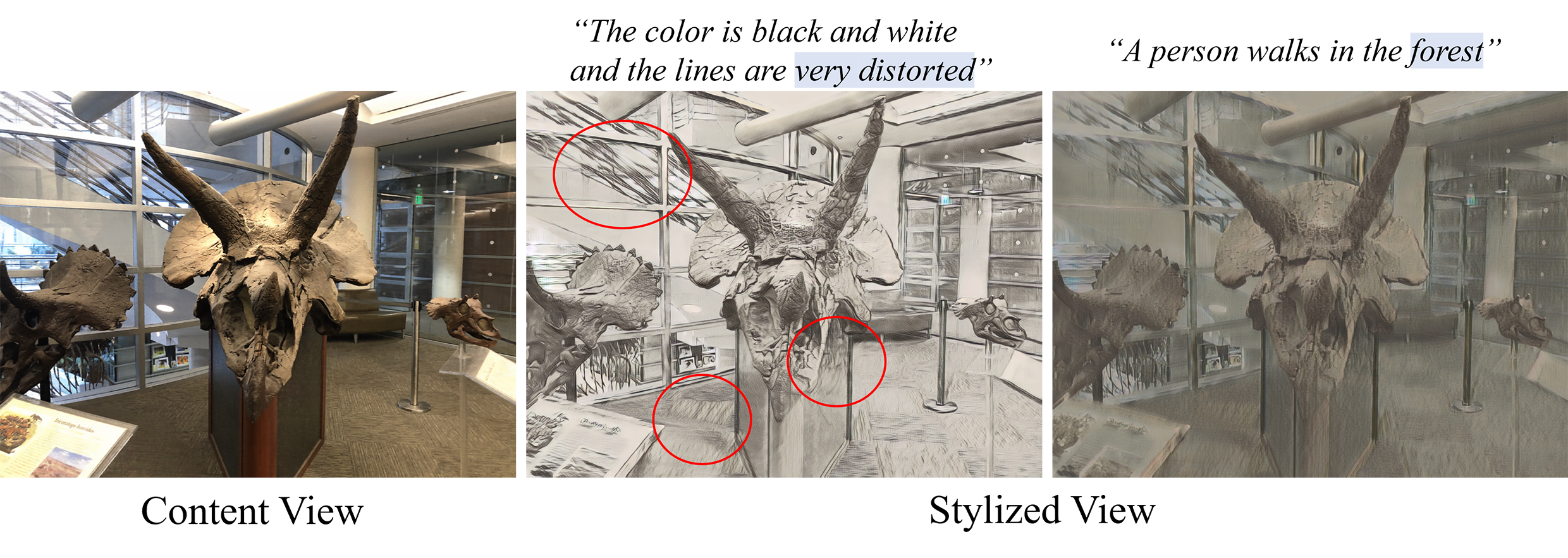}
  \caption{\textbf{Limitation.} While the red circle contains textual style information, the resultant overall outcome is unsatisfactory.}
  \label{fig:limitation}
\end{figure}

\subsection{Limitations} 
Our model benefits from CLIP, while also facing certain limitations imposed by it. For example, as shown in \Cref{fig:limitation}, we were not able to migrate the blue text style well because we used CLIP directly without making any changes to it. Our work currently only focuses on artistic style transfer and does not have creative capabilities based on generative models. We expect future work to successfully address them. Additionally, our local style transfer currently only works in face-forwarding scenes.

\section{Conclusion}
In this work, we introduce ConRF, a novel approach for achieving zero-shot stylization using either image or text as a reference. ConRF is mapping the CLIP features space to the VGG style space, enabling style transfer within the feature space of the scene. It can generate high-quality stylization novel views. Furthermore, we utilize a 3D volume to create local transfer effects, offering potential applications in artistic 3D design.

%
%
\bibliographystyle{splncs04}
\bibliography{main}

\begin{thebibliography}{10}
\providecommand{\url}[1]{\texttt{#1}}
\providecommand{\urlprefix}{URL }
\providecommand{\doi}[1]{https://doi.org/#1}

\bibitem{barron2021mip}
Barron, J.T., Mildenhall, B., Tancik, M., Hedman, P., Martin-Brualla, R., Srinivasan, P.P.: Mip-nerf: A multiscale representation for anti-aliasing neural radiance fields. In: Proceedings of the IEEE/CVF International Conference on Computer Vision. pp. 5855--5864 (2021)

\bibitem{chen2017coherent}
Chen, D., Liao, J., Yuan, L., Yu, N., Hua, G.: Coherent online video style transfer. In: Proceedings of the IEEE International Conference on Computer Vision. pp. 1105--1114 (2017)

\bibitem{chentestnerf}
Chen, J., Ji, B., Zhang, Z., Chu, T., Zuo, Z., Zhao, L., Xing, W., Lu, D.: Testnerf: Text-driven 3d style transfer via cross-modal learning

\bibitem{chen2022upst}
Chen, Y., Yuan, Q., Li, Z., Liu, Y., Wang, W., Xie, C., Wen, X., Yu, Q.: Upst-nerf: Universal photorealistic style transfer of neural radiance fields for 3d scene. arXiv preprint arXiv:2208.07059  (2022)

\bibitem{chiang2022stylizing}
Chiang, P.Z., Tsai, M.S., Tseng, H.Y., Lai, W.S., Chiu, W.C.: Stylizing 3d scene via implicit representation and hypernetwork. In: Proceedings of the IEEE/CVF Winter Conference on Applications of Computer Vision. pp. 1475--1484 (2022)

\bibitem{deng2022stytr2}
Deng, Y., Tang, F., Dong, W., Ma, C., Pan, X., Wang, L., Xu, C.: Stytr2: Image style transfer with transformers. In: Proceedings of the IEEE/CVF conference on computer vision and pattern recognition. pp. 11326--11336 (2022)

\bibitem{deng2020arbitrary}
Deng, Y., Tang, F., Dong, W., Sun, W., Huang, F., Xu, C.: Arbitrary style transfer via multi-adaptation network. In: Proceedings of the 28th ACM international conference on multimedia. pp. 2719--2727 (2020)

\bibitem{duan2023dynamic}
Duan, H., Long, Y., Wang, S., Zhang, H., Willcocks, C.G., Shao, L.: Dynamic unary convolution in transformers. IEEE Transactions on Pattern Analysis and Machine Intelligence  (2023)

\bibitem{painterbynumbers}
small~yellow duck, W.K.: Painter by numbers (2016), \url{https://kaggle.com/competitions/painter-by-numbers}

\bibitem{fan2022unified}
Fan, Z., Jiang, Y., Wang, P., Gong, X., Xu, D., Wang, Z.: Unified implicit neural stylization. In: European Conference on Computer Vision. pp. 636--654. Springer (2022)

\bibitem{fu2022language}
Fu, T.J., Wang, X.E., Wang, W.Y.: Language-driven artistic style transfer. In: European Conference on Computer Vision. pp. 717--734. Springer (2022)

\bibitem{gatys2015neural}
Gatys, L.A., Ecker, A.S., Bethge, M.: A neural algorithm of artistic style. arXiv preprint arXiv:1508.06576  (2015)

\bibitem{gatys2016image}
Gatys, L.A., Ecker, A.S., Bethge, M.: Image style transfer using convolutional neural networks. In: Proceedings of the IEEE conference on computer vision and pattern recognition. pp. 2414--2423 (2016)

\bibitem{instructnerf2023}
Haque, A., Tancik, M., Efros, A., Holynski, A., Kanazawa, A.: Instruct-nerf2nerf: Editing 3d scenes with instructions. In: Proceedings of the IEEE/CVF International Conference on Computer Vision (2023)

\bibitem{hertzmann1998painterly}
Hertzmann, A.: Painterly rendering with curved brush strokes of multiple sizes. In: Proceedings of the 25th annual conference on Computer graphics and interactive techniques. pp. 453--460 (1998)

\bibitem{10.1145/383259.383295}
Hertzmann, A., Jacobs, C.E., Oliver, N., Curless, B., Salesin, D.H.: Image analogies. In: Proceedings of the 28th Annual Conference on Computer Graphics and Interactive Techniques. p. 327–340. SIGGRAPH '01, Association for Computing Machinery, New York, NY, USA (2001). \doi{10.1145/383259.383295}, \url{https://doi.org/10.1145/383259.383295}

\bibitem{huang2017real}
Huang, H., Wang, H., Luo, W., Ma, L., Jiang, W., Zhu, X., Li, Z., Liu, W.: Real-time neural style transfer for videos. In: Proceedings of the IEEE conference on computer vision and pattern recognition. pp. 783--791 (2017)

\bibitem{huang2021learning}
Huang, H.P., Tseng, H.Y., Saini, S., Singh, M., Yang, M.H.: Learning to stylize novel views. In: Proceedings of the IEEE/CVF International Conference on Computer Vision. pp. 13869--13878 (2021)

\bibitem{huang2017arbitrary}
Huang, X., Belongie, S.: Arbitrary style transfer in real-time with adaptive instance normalization. In: Proceedings of the IEEE international conference on computer vision. pp. 1501--1510 (2017)

\bibitem{huang2022stylizednerf}
Huang, Y.H., He, Y., Yuan, Y.J., Lai, Y.K., Gao, L.: Stylizednerf: consistent 3d scene stylization as stylized nerf via 2d-3d mutual learning. In: Proceedings of the IEEE/CVF Conference on Computer Vision and Pattern Recognition. pp. 18342--18352 (2022)

\bibitem{johnson2016perceptual}
Johnson, J., Alahi, A., Fei-Fei, L.: Perceptual losses for real-time style transfer and super-resolution. In: Computer Vision--ECCV 2016: 14th European Conference, Amsterdam, The Netherlands, October 11-14, 2016, Proceedings, Part II 14. pp. 694--711. Springer (2016)

\bibitem{kamata2023instruct}
Kamata, H., Sakuma, Y., Hayakawa, A., Ishii, M., Narihira, T.: Instruct 3d-to-3d: Text instruction guided 3d-to-3d conversion. arXiv preprint arXiv:2303.15780  (2023)

\bibitem{kingma2014adam}
Kingma, D.P., Ba, J.: Adam: A method for stochastic optimization. arXiv preprint arXiv:1412.6980  (2014)

\bibitem{kwon2022clipstyler}
Kwon, G., Ye, J.C.: Clipstyler: Image style transfer with a single text condition. In: Proceedings of the IEEE/CVF Conference on Computer Vision and Pattern Recognition. pp. 18062--18071 (2022)

\bibitem{li2019learning}
Li, X., Liu, S., Kautz, J., Yang, M.H.: Learning linear transformations for fast image and video style transfer. In: Proceedings of the IEEE/CVF Conference on Computer Vision and Pattern Recognition. pp. 3809--3817 (2019)

\bibitem{li2017universal}
Li, Y., Fang, C., Yang, J., Wang, Z., Lu, X., Yang, M.H.: Universal style transfer via feature transforms. Advances in neural information processing systems  \textbf{30} (2017)

\bibitem{liu2023stylerf}
Liu, K., Zhan, F., Chen, Y., Zhang, J., Yu, Y., El~Saddik, A., Lu, S., Xing, E.P.: Stylerf: Zero-shot 3d style transfer of neural radiance fields. In: Proceedings of the IEEE/CVF Conference on Computer Vision and Pattern Recognition. pp. 8338--8348 (2023)

\bibitem{liu2020neural}
Liu, L., Gu, J., Zaw~Lin, K., Chua, T.S., Theobalt, C.: Neural sparse voxel fields. Advances in Neural Information Processing Systems  \textbf{33},  15651--15663 (2020)

\bibitem{liu2021adaattn}
Liu, S., Lin, T., He, D., Li, F., Wang, M., Li, X., Sun, Z., Li, Q., Ding, E.: Adaattn: Revisit attention mechanism in arbitrary neural style transfer. In: Proceedings of the IEEE/CVF international conference on computer vision. pp. 6649--6658 (2021)

\bibitem{luan2017deep}
Luan, F., Paris, S., Shechtman, E., Bala, K.: Deep photo style transfer. In: Proceedings of the IEEE conference on computer vision and pattern recognition. pp. 4990--4998 (2017)

\bibitem{miao2024ctnerf}
Miao, X., Bai, Y., Duan, H., Huang, Y., Wan, F., Long, Y., Zheng, Y.: Ctnerf: Cross-time transformer for dynamic neural radiance field from monocular video. arXiv preprint arXiv:2401.04861  (2024)

\bibitem{miao2023ds}
Miao, X., Bai, Y., Duan, H., Huang, Y., Wan, F., Xu, X., Long, Y., Zheng, Y.: Ds-depth: Dynamic and static depth estimation via a fusion cost volume. IEEE Transactions on Circuits and Systems for Video Technology  (2023)

\bibitem{mildenhall2022nerf}
Mildenhall, B., Hedman, P., Martin-Brualla, R., Srinivasan, P.P., Barron, J.T.: Nerf in the dark: High dynamic range view synthesis from noisy raw images. In: Proceedings of the IEEE/CVF Conference on Computer Vision and Pattern Recognition. pp. 16190--16199 (2022)

\bibitem{mildenhall2019local}
Mildenhall, B., Srinivasan, P.P., Ortiz-Cayon, R., Kalantari, N.K., Ramamoorthi, R., Ng, R., Kar, A.: Local light field fusion: Practical view synthesis with prescriptive sampling guidelines. ACM Transactions on Graphics (TOG)  \textbf{38}(4),  1--14 (2019)

\bibitem{mildenhall2021nerf}
Mildenhall, B., Srinivasan, P.P., Tancik, M., Barron, J.T., Ramamoorthi, R., Ng, R.: Nerf: Representing scenes as neural radiance fields for view synthesis. Communications of the ACM  \textbf{65}(1),  99--106 (2021)

\bibitem{mu20223d}
Mu, F., Wang, J., Wu, Y., Li, Y.: 3d photo stylization: Learning to generate stylized novel views from a single image. In: Proceedings of the IEEE/CVF Conference on Computer Vision and Pattern Recognition. pp. 16273--16282 (2022)

\bibitem{nguyen2022snerf}
Nguyen-Phuoc, T., Liu, F., Xiao, L.: Snerf: stylized neural implicit representations for 3d scenes. arXiv preprint arXiv:2207.02363  (2022)

\bibitem{park2019arbitrary}
Park, D.Y., Lee, K.H.: Arbitrary style transfer with style-attentional networks. In: proceedings of the IEEE/CVF conference on computer vision and pattern recognition. pp. 5880--5888 (2019)

\bibitem{radford2021learning}
Radford, A., Kim, J.W., Hallacy, C., Ramesh, A., Goh, G., Agarwal, S., Sastry, G., Askell, A., Mishkin, P., Clark, J., et~al.: Learning transferable visual models from natural language supervision. In: International conference on machine learning. pp. 8748--8763. PMLR (2021)

\bibitem{risser2017stable}
Risser, E., Wilmot, P., Barnes, C.: Stable and controllable neural texture synthesis and style transfer using histogram losses. arXiv preprint arXiv:1701.08893  (2017)

\bibitem{ruder2018artistic}
Ruder, M., Dosovitskiy, A., Brox, T.: Artistic style transfer for videos and spherical images. International Journal of Computer Vision  \textbf{126}(11),  1199--1219 (2018)

\bibitem{sheng2018avatar}
Sheng, L., Lin, Z., Shao, J., Wang, X.: Avatar-net: Multi-scale zero-shot style transfer by feature decoration. In: Proceedings of the IEEE conference on computer vision and pattern recognition. pp. 8242--8250 (2018)

\bibitem{simonyan2014very}
Simonyan, K., Zisserman, A.: Very deep convolutional networks for large-scale image recognition. arXiv preprint arXiv:1409.1556  (2014)

\bibitem{teed2020raft}
Teed, Z., Deng, J.: Raft: Recurrent all-pairs field transforms for optical flow. In: Computer Vision--ECCV 2020: 16th European Conference, Glasgow, UK, August 23--28, 2020, Proceedings, Part II 16. pp. 402--419. Springer (2020)

\bibitem{ulyanov2016texture}
Ulyanov, D., Lebedev, V., Vedaldi, A., Lempitsky, V.: Texture networks: Feed-forward synthesis of textures and stylized images. arXiv preprint arXiv:1603.03417  (2016)

\bibitem{wang2022clip}
Wang, C., Chai, M., He, M., Chen, D., Liao, J.: Clip-nerf: Text-and-image driven manipulation of neural radiance fields. In: Proceedings of the IEEE/CVF Conference on Computer Vision and Pattern Recognition. pp. 3835--3844 (2022)

\bibitem{wang2020consistent}
Wang, W., Xu, J., Zhang, L., Wang, Y., Liu, J.: Consistent video style transfer via compound regularization. In: Proceedings of the AAAI conference on artificial intelligence. vol.~34, pp. 12233--12240 (2020)

\bibitem{wu2021styleformer}
Wu, X., Hu, Z., Sheng, L., Xu, D.: Styleformer: Real-time arbitrary style transfer via parametric style composition. In: Proceedings of the IEEE/CVF International Conference on Computer Vision. pp. 14618--14627 (2021)

\bibitem{wu2022ccpl}
Wu, Z., Zhu, Z., Du, J., Bai, X.: Ccpl: contrastive coherence preserving loss for versatile style transfer. In: European Conference on Computer Vision. pp. 189--206. Springer (2022)

\bibitem{xian2021space}
Xian, W., Huang, J.B., Kopf, J., Kim, C.: Space-time neural irradiance fields for free-viewpoint video. In: Proceedings of the IEEE/CVF Conference on Computer Vision and Pattern Recognition. pp. 9421--9431 (2021)

\bibitem{xiangli2022bungeenerf}
Xiangli, Y., Xu, L., Pan, X., Zhao, N., Rao, A., Theobalt, C., Dai, B., Lin, D.: Bungeenerf: Progressive neural radiance field for extreme multi-scale scene rendering. In: Computer Vision--ECCV 2022: 17th European Conference, Tel Aviv, Israel, October 23--27, 2022, Proceedings, Part XXXII. pp. 106--122. Springer (2022)

\bibitem{xu2022point}
Xu, Q., Xu, Z., Philip, J., Bi, S., Shu, Z., Sunkavalli, K., Neumann, U.: Point-nerf: Point-based neural radiance fields. In: Proceedings of the IEEE/CVF Conference on Computer Vision and Pattern Recognition. pp. 5438--5448 (2022)

\bibitem{yu2021pixelnerf}
Yu, A., Ye, V., Tancik, M., Kanazawa, A.: pixelnerf: Neural radiance fields from one or few images. In: Proceedings of the IEEE/CVF Conference on Computer Vision and Pattern Recognition. pp. 4578--4587 (2021)

\bibitem{zhang2022arf}
Zhang, K., Kolkin, N., Bi, S., Luan, F., Xu, Z., Shechtman, E., Snavely, N.: Arf: Artistic radiance fields. In: European Conference on Computer Vision. pp. 717--733. Springer (2022)

\bibitem{zhang2018unreasonable}
Zhang, R., Isola, P., Efros, A.A., Shechtman, E., Wang, O.: The unreasonable effectiveness of deep features as a perceptual metric. In: Proceedings of the IEEE conference on computer vision and pattern recognition. pp. 586--595 (2018)

\bibitem{zhang2023ref}
Zhang, Y., He, Z., Xing, J., Yao, X., Jia, J.: Ref-npr: Reference-based non-photorealistic radiance fields for controllable scene stylization. In: Proceedings of the IEEE/CVF Conference on Computer Vision and Pattern Recognition. pp. 4242--4251 (2023)

\bibitem{zhang2023transforming}
Zhang, Z., Liu, Y., Han, C., Pan, Y., Guo, T., Yao, T.: Transforming radiance field with lipschitz network for photorealistic 3d scene stylization. In: Proceedings of the IEEE/CVF Conference on Computer Vision and Pattern Recognition. pp. 20712--20721 (2023)

\end{thebibliography}
\clearpage

\section{Supplementary}

\subsection{Implementation details}
\begin{figure}
  \centering
  \includegraphics[width=0.4\linewidth]{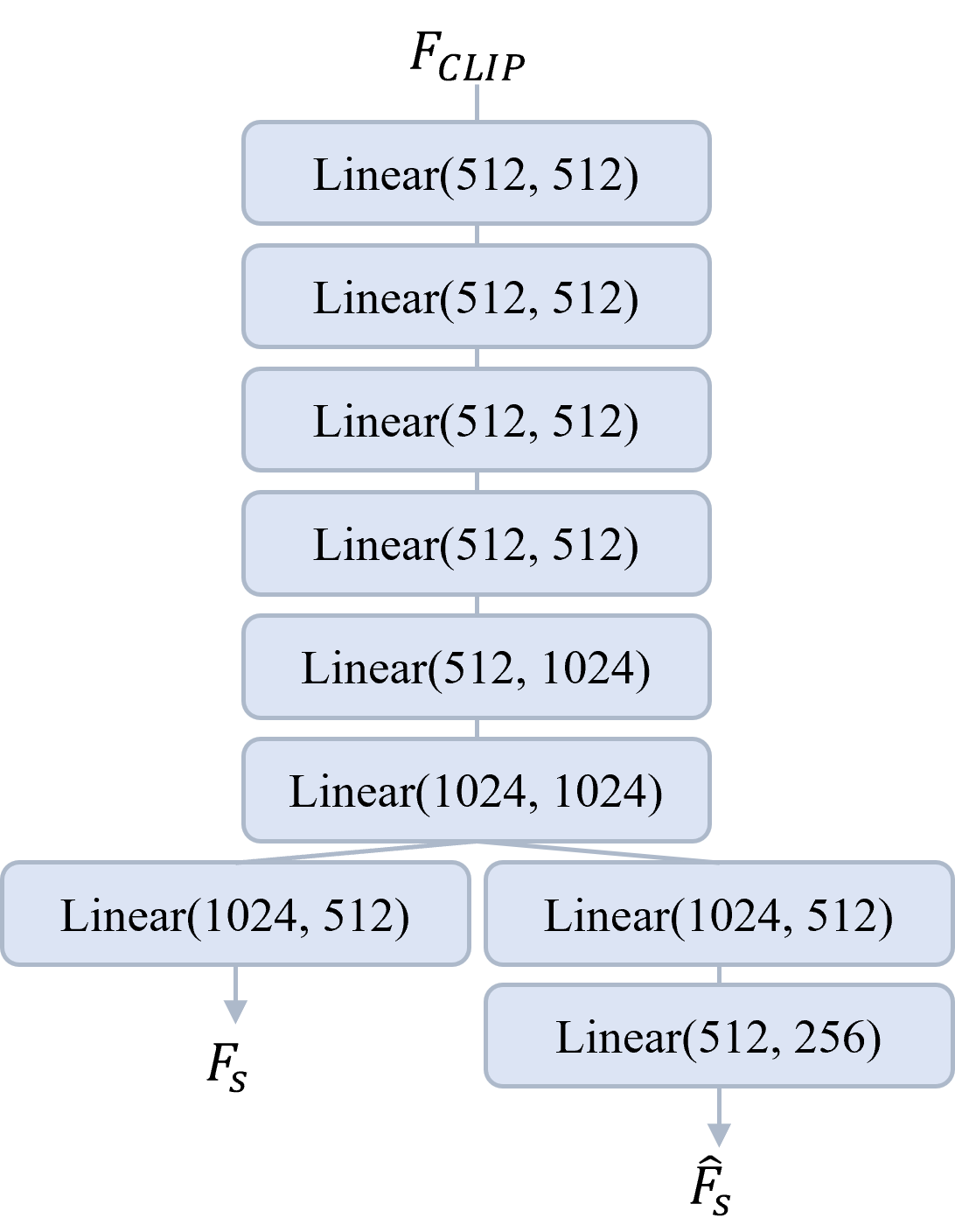}
  \caption{\textbf{Mapping network architecture.} $F_s$ represents the style feature, which consists of the mean and standard deviation obtained through mapping the CLIP feature ($F_\text{CLIP}$). Meanwhile, $\hat{F_s}$ denotes the stylized feature.}
  \label{fig:architecture}
\end{figure}

Our model is trained on a single RTX 3090 GPU using approximately 80,000 samples from the WiKiArt dataset \cite{painterbynumbers} as style images. During the training phase, the Adam optimizer \cite{kingma2014adam} is employed with a learning rate of 1e-4. We assign the loss weights for style and content as 20 and 1, respectively, while all other loss weights are maintained at 1. For incorporating CLIP \cite{radford2021learning}, we utilize the pre-trained ViT-B/32 model.
\paragraph{Mapping network architecture}
We present the architecture of the mapping network, the details please see \cref{fig:architecture}.

\begin{table}
\resizebox{\textwidth}{!}{
\begin{tabular}{c|c|c|c|c|c}
\toprule
Method    & Type & Need Retrain & Local Transfer & Guided with Image & Guided with Text \\ \hline
ARF  \cite{zhang2022arf}     &      S        &     \Checkmark        &     \XSolidBrush      &          \Checkmark        &             \XSolidBrush                               \\
SNeRF \cite{nguyen2022snerf}    &      S        &    \Checkmark         &   \XSolidBrush        &         \Checkmark        &             \XSolidBrush                            \\
Ref-NPR \cite{zhang2023ref}   &     M         &      \Checkmark       &   \XSolidBrush        &    \Checkmark        &             \XSolidBrush                              \\
StyleRF \cite{liu2023stylerf}  &      A        &    \XSolidBrush         &    \Checkmark       &        \Checkmark        &             \XSolidBrush                       \\
CLIP-NeRF \cite{wang2022clip} &       A       &  \XSolidBrush           &    \XSolidBrush       &       \Checkmark         &           \Checkmark                         \\ \hline
Our       &      A        &    \XSolidBrush         &   \Checkmark        &     \Checkmark          &            \Checkmark                        \\ \bottomrule
\end{tabular}
}
\caption{\textbf{Comparison of 3D style transfer methods.}  We compared other types of 3D art style transfer methods. S=Single-Style transfer, M=Multi-Style transfer, and A=Arbitrary Style transfer.}
\label{tab:2}
\end{table}

\subsection{User study}
We conduct a user study, following previous research \cite{liu2023stylerf}, where we compare our approach to a 3D style transfer baseline. We engage 25 participants with diverse demographic backgrounds, including different professions, age ranges, and ethnicities, to ensure a broad perspective. Each participant is shown a series of stylizations that include original scene videos, their stylized counterparts, and videos processed using both our ConRF method and the baseline technique.

Participants are tasked with two evaluations: identifying the stylized video that most accurately reflects the style of a given image, and selecting the video that exhibits the best multi-view consistency. For a comprehensive analysis, we prepare 24 unique scene-style pairings, which we then randomly distribute into 6 categories. Each participant evaluates one category, providing us with a diverse set of opinions on the effectiveness of our approach. The aggregated findings from this study are detailed in Table 2, as referenced in our report.

\begin{table}
\centering
\begin{tabular}{ccccc}
\toprule
Methods & Stylization Score &  Consistency Score \\ \hline
Ref-NPR \cite{zhang2023ref} & 0.473                                     & 0.766\\
SNeRF \cite{nguyen2022snerf}   & \color{blue}{0.526}                                     & 0.696\\
ARF \cite{zhang2022arf}    & 0.450                                     & 0.795\\
StyleRF \cite{liu2023stylerf} & 0.510                                     & \textcolor{blue}{0.800}\\
\textbf{Our}     & \textcolor{red}{0.693}                                     & \textcolor{red}{0.843}\\ \hline 
$\text{CLIPStyler}^*$ \cite{kwon2022clipstyler}
 & \textcolor{blue}{0.605}                                     & 0.333\\
$\text{CLIP-NeRF}^*$ \cite{wang2022clip}
   & 0.560                                     & \textcolor{red}{0.920}\\
$\text{\textbf{Our}}^*$     & \textcolor{red}{0.620}                                     & \textcolor{blue}{0.880}\\
\bottomrule
\end{tabular}
\caption{\textbf{User study.}  We compare ConRF with the state-of-the-art 3D style transfer baseline. ($*$) is using a text prompt to style transfer. The best score is \textcolor{red}{red}, and the second score is \textcolor{blue}{blue}.}
\label{tab:2}
\end{table}

\begin{figure*}
  \centering
  \includegraphics[width=0.8\linewidth]{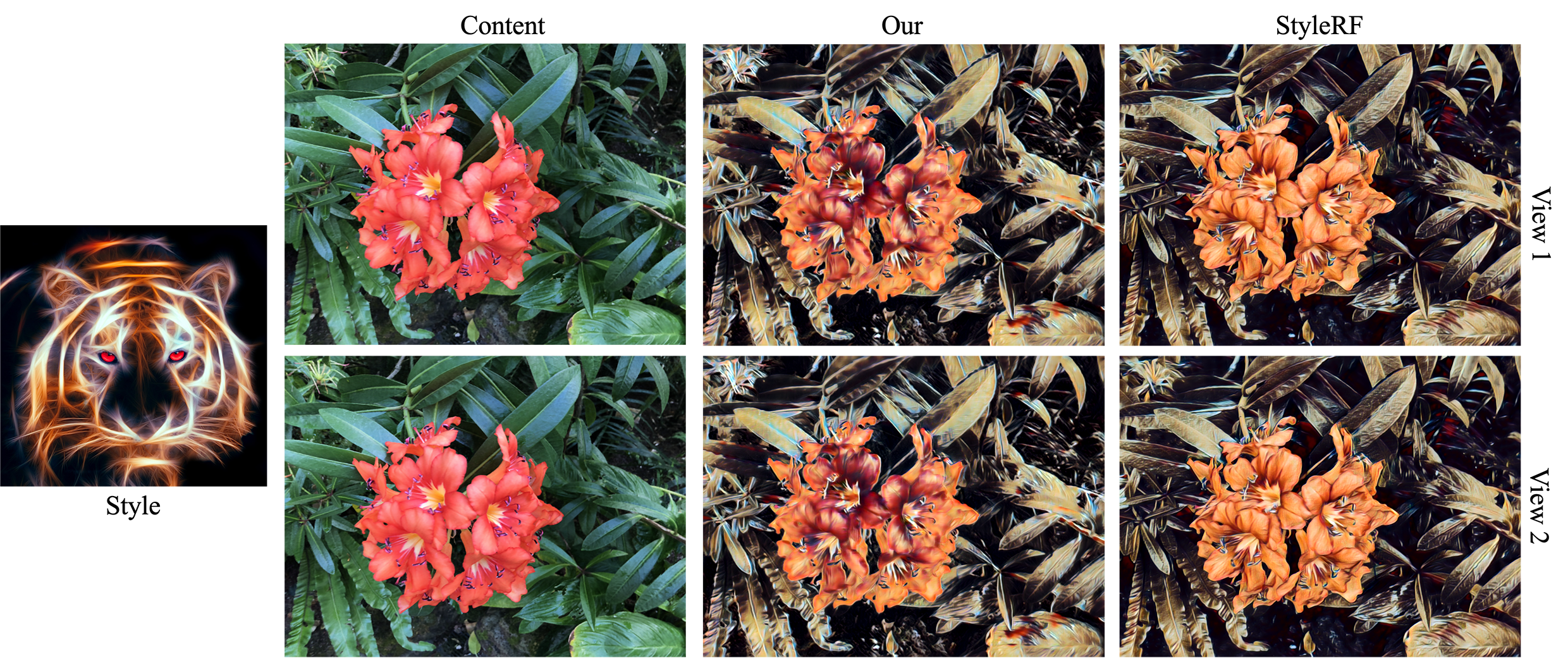}
  \caption{\textbf{Comparison with StyleRF \cite{liu2023stylerf} with different views on LLFF.} Here, we present the flower scene style transfer results.}
  \label{fig:2views}
\end{figure*}

\subsection{Additional Experiment}
We present more qualitative results here and video results in the supplementary material.
\subsubsection{Qualitative results}
\begin{figure*}
  \centering
  \includegraphics[width=0.9\linewidth]{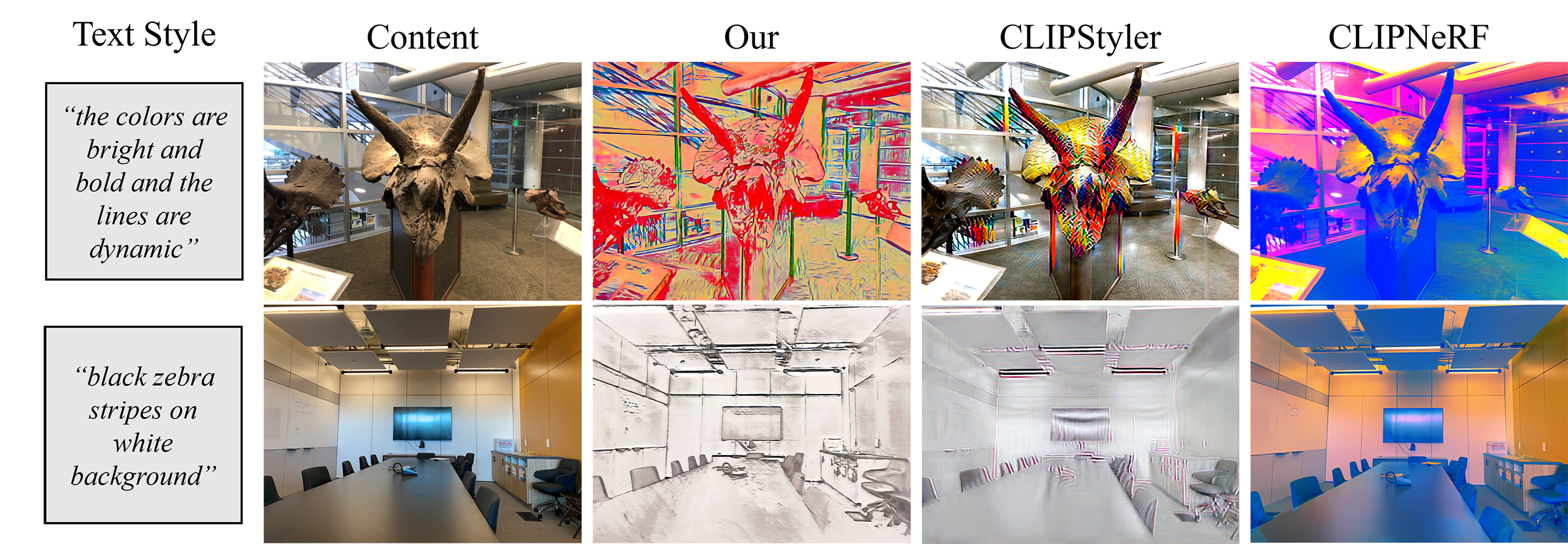}
  \caption{\textbf{Comparison with two SOTA style transfer methods using text prompt on LLFF.} Here, we present the horns and room scenes style transfer results.}
  \label{fig:supp_text_1}
\end{figure*}

\paragraph{Text prompt style transfer comparison on LLFF database} As shown in \cref{fig:supp_text_1}, we additionally perform evaluations on the LLFF \cite{mildenhall2021nerf}.

\begin{figure*}
  \centering
  \includegraphics[width=0.8\linewidth]{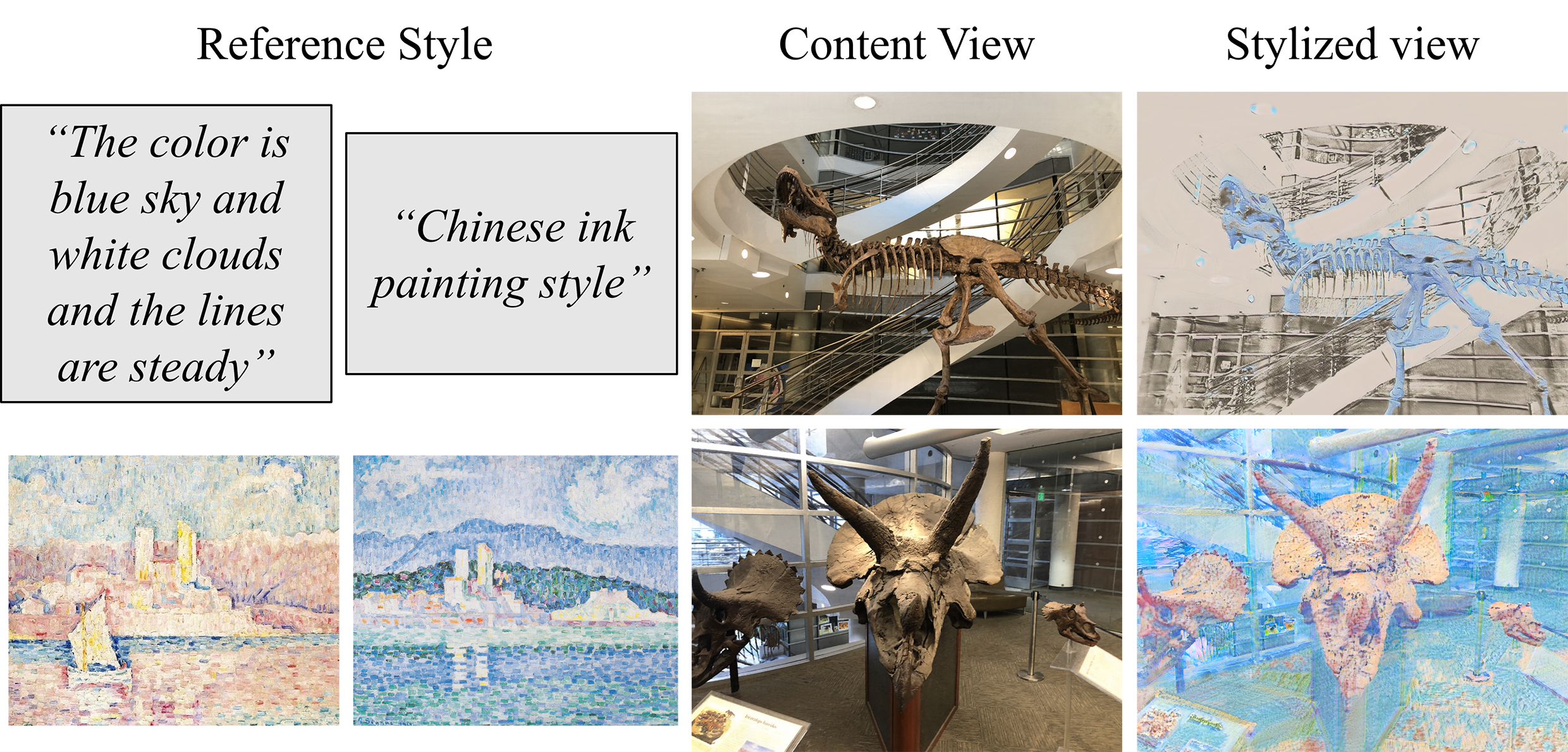}
  \caption{\textbf{Local style transfer.} In the main text, we present the transfer results of the text-image combination, where we show the transfer results of text-text and image-image.}
  \label{fig:supp_3}
\end{figure*}

\paragraph{Local style transfer quantitative results} 
As outlined in the main text, our model is versatile, supporting text-image, text-text, and image-image combinations to generate diverse stylization outcomes. The text-text and image-image combinations, which demonstrate this flexibility, are exemplified in Figure 3 of the supplementary materials (\cref{fig:supp_3}).

\begin{figure*}
  \centering
  \includegraphics[width=0.8\linewidth]{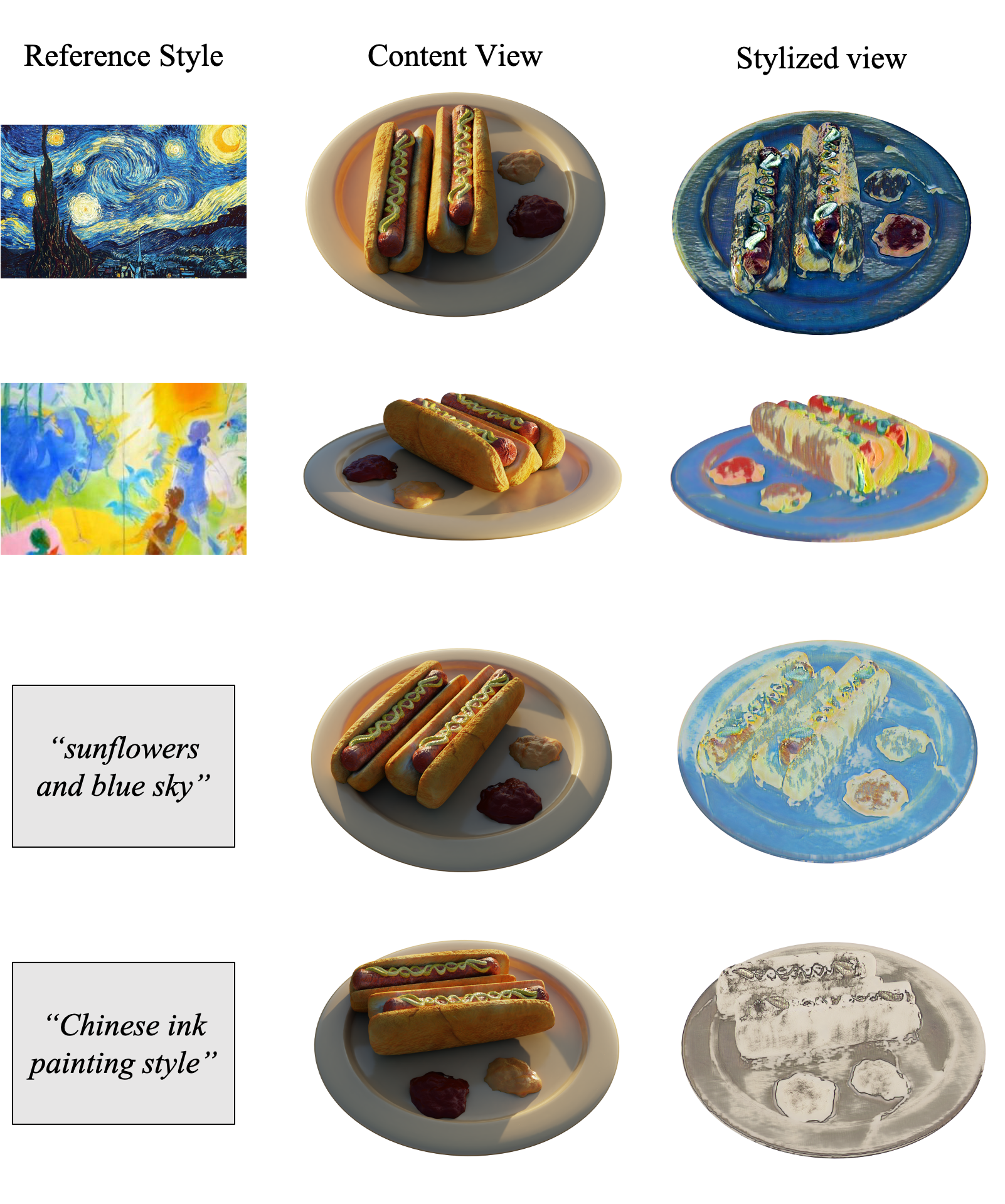}
  \caption{\textbf{Style transfer result on Synthetic NeRF dataset.} We present the text and image style transfer results of the hotdog scene.}
  \label{fig:supp_2}
\end{figure*}

\begin{figure*}
  \centering
  \includegraphics[width=0.8\linewidth]{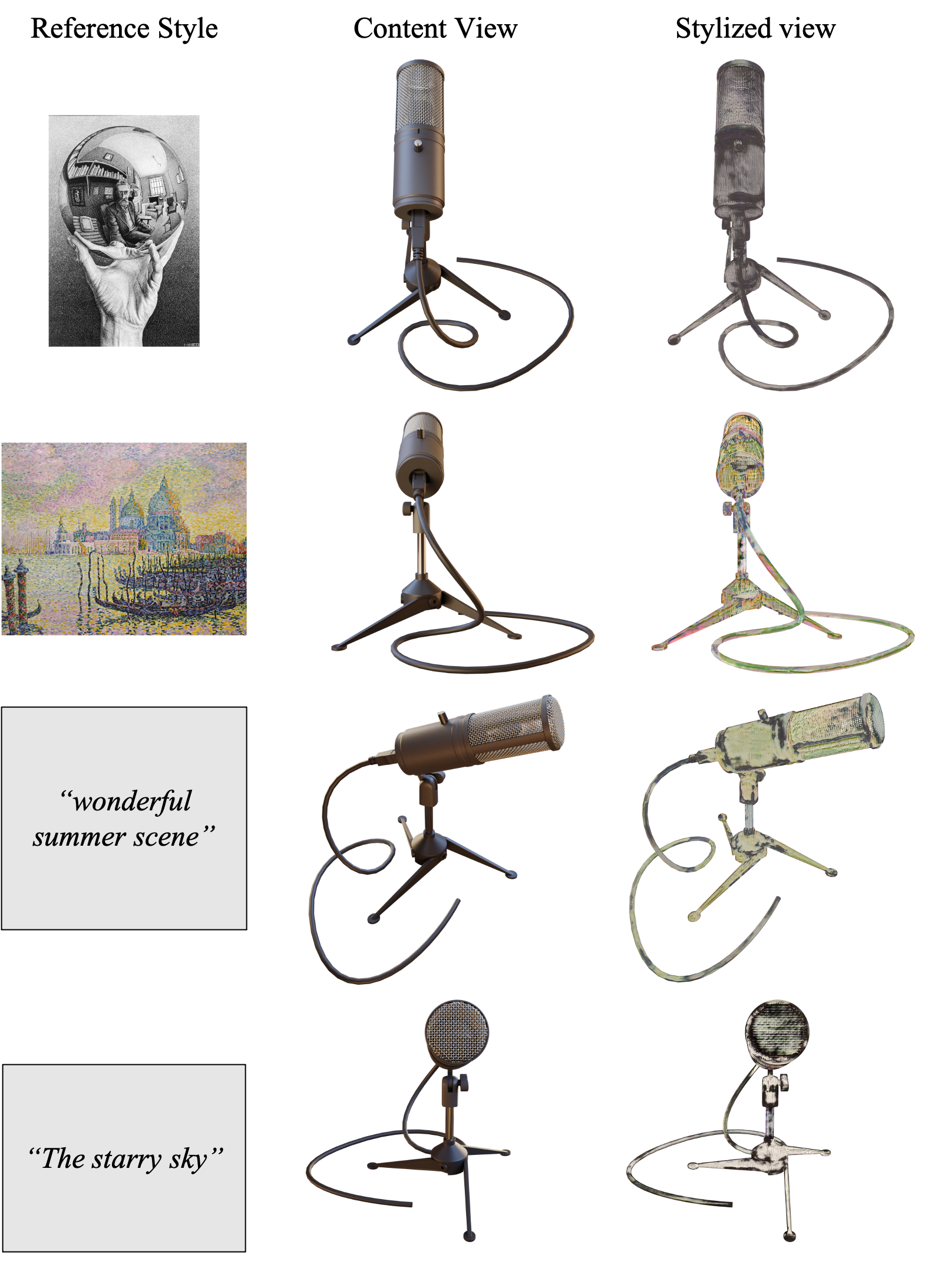}
  \caption{\textbf{Style transfer result on Synthetic NeRF dataset.} We present the text and image style transfer results of the mic scene.}
  \label{fig:supp_4}
\end{figure*}

\paragraph{More quantitative results} To demonstrate the performance of our method, as shown in the \cref{fig:2views}, \cref{fig:supp_2}, and \cref{fig:supp_4}, here we provide more visualization results. 


\end{document}